\documentclass[preprint,12pt,authoryear]{elsarticle}

\usepackage{nicefrac}       
\usepackage{microtype}      
\usepackage{graphicx}
\usepackage{subcaption}

\usepackage{bm}
\usepackage{amssymb}
\usepackage{mathtools}
\usepackage[mathscr]{eucal}
\usepackage{xspace}
\usepackage{color}

\def\w{{\bf w}}
\def\x{{\bf x}}
\def\xa{{\bm\phi}}
\def\h{{\bm\theta}}
\def\e{{\bf e}}

\def\la{({$\lambda$})~}
\def\Re{{\mathbb{R}}}

\newcommand{\EE}[2]{\mathbb{E}_{#1\!\!}\left[#2\right]}

\newcommand{\CEE}[3]{\EE{#1}{{#2}~\middle\vert~{#3}}}

\def\CEpi#1#2{\CEE{\pi}{#1}{#2}}
\def\CP#1#2{\Pr{#1\mid#2}}
\def\Pr#1{{{\rm Pr\!}\left\{#1\right\}}}
\def\eqrf#1{\eqref{eq:#1}}
\def\tr{{^\top}}
\def\S{\mathscr{S}}
\def\A{\mathscr{A}}
\def\T{\mathscr{T}}
\def\N{\mathbb{N}}
\def\O{\mathscr{O}}
\def\R{{\cal R}}
\def\ra{\rightarrow}

\def\E#1{\EE{\,}{#1}}
\def\={\! = \!}
\def\d={\!\doteq\!}
\def\v{{\hat v}}
\def\bw#1{\overline{w}^{\hspace{0.5pt}#1}}
\def\la{($\lambda$)\xspace}
\def\UWT{\mbox{\sf UWT}\xspace}
\def\r{{\hat r}}
\def\n{{\bf\hat n}}
\def\W{{\bf W}}

\journal{Artificial Intelligence}

\begin{document}

\begin{frontmatter}

\title{Reward-Respecting Subtasks\\for Model-Based Reinforcement Learning}

\author[deepmind,uofa,amii,cifar]{Richard S. Sutton}

\affiliation[deepmind]{organization={DeepMind},
            city={Edmonton},
            state={Alberta},
            country={Canada}}

\affiliation[uofa]{organization={University of Alberta},
            city={Edmonton},
            state={Alberta},
            country={Canada}}
            
\affiliation[amii]{organization={Alberta Machine Intelligence Institute (Amii)},
            city={Edmonton},
            state={Alberta},
            country={Canada}}

\affiliation[cifar]{organization={Canada CIFAR AI Chair},
            country = {Canada}}

\author[deepmind,uofa,amii,cifar]{Marlos C. Machado\corref{cor1}}
\ead{machado@ualberta.ca}
\cortext[cor1]{Corresponding author}
\author[deepmind]{G. Zacharias Holland}
\author[deepmind]{David Szepesvari}
\author[deepmind]{Finbarr Timbers}
\author[deepmind]{Brian Tanner}
\author[deepmind,uofa,amii,cifar]{Adam White}

\begin{abstract}
To achieve the ambitious goals of artificial intelligence, reinforcement learning must include planning with a model of the world that is abstract in state and time. Deep learning has made progress with state abstraction, but temporal abstraction has rarely been used, despite extensively developed theory based on the options framework. One reason for this is that the space of possible options is immense, and the methods previously proposed for option discovery do not take into account how the option models will be used in planning. Options are typically discovered by posing subsidiary tasks, such as reaching a bottleneck state or maximizing the cumulative sum of a sensory signal other than reward. Each subtask is solved to produce an option, and then a model of the option is learned and made available to the planning process. In most previous work, the subtasks ignore the reward on the original problem, whereas we propose subtasks that use the original reward plus a bonus based on a feature of the state at the time the option terminates. We show that option models obtained from such reward-respecting subtasks are much more likely to be useful in planning than eigenoptions, shortest path options based on bottleneck states, or reward-respecting options generated by the option-critic. Reward respecting subtasks strongly constrain the space of options and thereby also provide a partial solution to the problem of option discovery. Finally, we show how values, policies, options, and models can all be learned online and off-policy using standard algorithms and general value functions.
\end{abstract}




\end{frontmatter}


\section{The challenge of discovering temporal abstractions}
\label{Introduction}

A major goal of artificial intelligence (AI) is to understand how an AI agent can obtain and reason with a high-level model of the world. The interaction between AI agent and world consists entirely of low-level, fleeting signals such as pixel values and motor torques, yet from this the agent must produce a model of the world that supports reasoning about high-level things such as which object to pick up, whether to walk to work or drive, or which country to vacation in. How abstractions can be obtained to bridge the gap between the low and high levels in planning is a recurring theme in the history of AI (e.g., Sacerdoti, 1974; Knoblock, 1994; Konidaris, Kaelbling \& Lozano-Perez, 2018).

In the reinforcement learning (RL) approach to AI, the issue of abstraction in planning arises in model-based RL, in which an RL agent learns a model of the transition dynamics of its environment and then converts that model into improvements in its policy and, commonly, in its approximate value function. This conversion process is planning and is typically computationally expensive and distributed over many time steps, or even performed offline. Learning and planning with a model may enable dramatically faster adaptation whenever the agent is long-lived, the environment is non-stationary, and much of the environment's transition dynamics is stable (as approximately modeled by the agent). 

For planning to be tractable on large problems, the RL agent's model must be abstract in state and time.  Abstraction in state is important because the original states of the world are too numerous to deal with individually, or may not be observable by the agent. In these cases, how the state should be constructed from observations is an important problem on which much work has been done with deep learning~(e.g., Hochreiter \& Schmidhuber, 1997; LeCun et al., 1998; Mnih et al., 2015) and other methods (e.g., Rivest \& Schapire, 1994; Littman et al., 2002; Jaeger, 2000).
We do not address state abstraction in this paper other than by allowing the agent's state representation to be a non-Markov feature vector.

This paper concerns how we should create and work with environment models that are abstract in \emph{time}. The most common way of formulating temporally-extended and temporally-variable ways of behaving in a reinforcement learning agent is as \emph{options}~(Sutton, Precup \& Singh, 1999), each of which comprises a way of behaving (a policy) and a way of stopping.
The appeal of options is that they are in some ways interchangeable with actions. Just as we can learn models of action's consequences and plan with those models, so we can learn and plan with models of options' effects.

There remains the critical question of where the options come from. A common approach to option discovery is to pose subsidiary tasks such as reaching a bottleneck state or maximizing the cumulative sum of a sensory signal other than reward. Given such subtasks, the agent can develop temporally abstract structure for its cognition by following a standard progression in which each subtask is solved to produce an option, the option's consequences are learned to produce a model, and the model is used in planning. We refer to this progression (SubTask, Option, Model, Planning) as the STOMP progression for the development of temporally-abstract cognitive structure. All the steps of the STOMP progression were described in the original paper on options (Sutton et al., 1999), and the progression has been used in several previous works~(e.g., Singh, Barto \& Chentanez, 2004; Sorg~\&~Singh, 2010; Silver \& Ciosek, 2012; Ring, 2021; Veeriah, 2022). The steps of the progression are often most simply described and implemented sequentially, as they are in this paper, but there is no reason that they cannot proceed simultaneously, in parallel. This is in fact the architecture envisioned in the long run.

The primary conceptual innovation of the current work is to introduce the notion of a \emph{reward-respecting subtask}, that is, of a subtask that optimizes the rewards of the original task until terminating in a state that is sometimes of high value. Reward-respecting subtasks contrast with commonly used subtasks, such as shortest path to bottleneck states~(e.g., McGovern \& Barto, 2001; Simsek \& Barto, 2004; Simsek, Wolfe \& Barto, 2005), pixel maximization~(Jaderberg et al., 2017), and diffusion maximization~(see~Machado et al., 2023), many of  which explicitly maximize the cumulative sum of a signal other than the reward of the original task.

\begin{figure}
\includegraphics[width=0.9\textwidth]{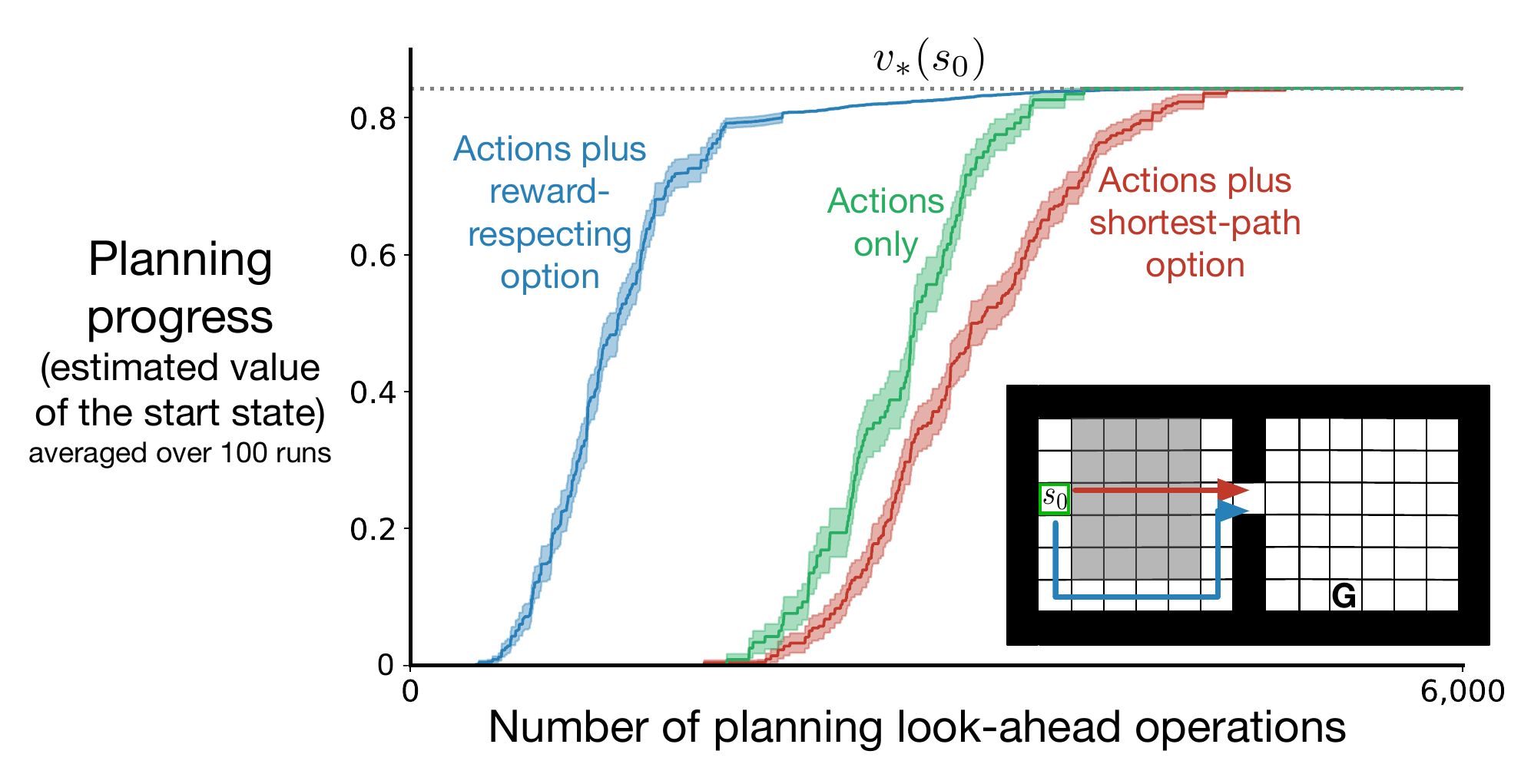}
\caption{\label{fig:exp1}
\textbf{Illustrative example} using the two-room gridworld (shown inset) contrasting planning with reward-respecting and shortest-path options for reaching the bottleneck state. In all cases the planner was given accurate models of the actions and options. Planning with the model of the reward-respecting option was much more efficient. The shading represents one standard error.}
\end{figure}

For example, consider the two-room gridworld shown inset in Figure~\ref{fig:exp1}, with a start state in one room, a terminal goal state in the other, and a bottleneck or \emph{hallway} state in-between. The usual four actions move the agent one cell up, down, right, or left, unless blocked by a wall. A reward of $+1$ is received on reaching the goal state, which ends the episode. Transitions ending in the gray region between the start and hallway states produce a reward of $-1$ per step, while all other transitions produce a reward of zero. The discount factor is $\gamma = 0.99$, so the optimal path from start to goal, traveling the roundabout path that avoids the field of negative reward, yields a return of $v_*(s_0)\=0.99^{17}\!\approx\!0.843$. The hallway state is a bottleneck and thus is a natural terminating subgoal for a subtask on this problem (McGovern \& Barto, 2001; Solway et al., 2014). With a reward-respecting subtask, the agent would learn a path to the hallway state that maximizes the reward along the way; in this gridworld it finds the option that goes down from the start state and \emph{around} the field of negative rewards. In contrast, solving the version of this subtask that does not take the reward into consideration would learn the option taking the shortest path from start to hallway, passing \emph{through} the field of negative rewards.

Which of these two options---\emph{reward respecting} or \emph{shortest path}---is the more useful when their models are learned and used in planning? Assuming optimal options and that
their models and those of the primitive actions are accurate, we can compare the progress of planning by value iteration (as detailed later in this paper) when augmented with models of each of the two options (Figure~\ref{fig:exp1}). 
In all cases, planning eventually found the optimal policy and the correct value of the start state. However, planning using the reward-respecting option was much faster than planning using the shortest-path option or planning using the primitive actions alone. Planning with the shortest-path option was actually \emph{slower} by this measure than planning with actions alone.
This could be because the planning with options required one additional look-ahead operation per state updated (four instead of five). (Planning with the reward-respecting option was more efficient \emph{despite} this disadvantage.) Or the poor performance of planning with the shortest-path option may just have been because that option was rarely part of a good trajectory. The shortest-path option passes through the field of negative reward, while the reward-respecting option follows the roundabout path hugging the bottom of the first room. The latter is much more likely to be part of the final optimal policy.

The next four sections detail the four steps of the STOMP progression in sequence. We formalize the concept of subtasks through the language of general value functions (GVFs) in Section 2. This perspective allows us to clearly introduce the notion of reward-respecting subtask. Moreover, it provides clarity in the option-discovery process by showing that different option-discovery methods differ only in how they define the GVFs' cumulants and stopping values. Given these subtasks, the subsequent sections detail option learning~(Section~\ref{sec:option_learning_feat_attainment}), model learning~(Section~\ref{sec:model_learning}), and planning~(Section~\ref{sec:planning_with_options}). In Section~\ref{sec:option_learning_feat_attainment} we introduce a general update procedure, \UWT, for learning both option and models through various temporal-difference errors, showing how the learning problems differ only in their targets. We further analyze the impact of different ways of defining subtasks in Section~\ref{sec:bonus-weights}. Our most important empirical results, comparing different methods for discovering options in a stochastic environment, are in the penultimate Section~\ref{sec:final_experiments}. 

\section{Reward-respecting subtasks}\label{sec:reward_respecting_subtasks}

In this section we define the agent--environment interaction, the main task, GVF subtasks, reward-respecting subtasks, reward-respecting subtasks of feature attainment, and the specific subtasks used in the illustrative example (Figure~\ref{fig:exp1}). 
Subtask specification is the first step in the STOMP progression for developing temporally-abstract cognitive structure.

We consider an agent interacting with its environment in a sequence of episodes, each beginning in environment state $S_0\d=s_0\in\S$ and ending in terminal state $S_L\d=\bot$\footnote{The symbol $\doteq$ is used in this paper to denote an equality that holds by definition rather than one that follows from previously established definitions.} at time step $L\in\N$. 
At time steps $t<L$, the agent selects an action $A_t \in \A$, and in response the environment emits a reward $R_{t+1} \in \R\subset\Re$ and transitions to a next state $S_{t + 1} \in \S+\bot$ with probability $p(s',r | s, a) \doteq \CP{S_{t+1}\=s', R_{t+1}\=r}{S_t\=s, A_t\=a}$, where $\A$, $\S$, and $\R$ are finite sets. Capitalized letters denote random variables that differ from  step to step and episode to episode; technically these should be indexed by an episode number, but we suppress that in our notation. 

The agent's main task is to find a policy  $\pi : \S \times \A \rightarrow [0,1]$ that maximizes the expected discounted sum of rewards 
$\E{R_1+\gamma R_2+\cdots +\gamma^{L-1}R_L}$ where $\gamma \in [0, 1)$ is the \emph{discount rate}, a parameter of the problem.
The value function $v_\pi:\S\ra\Re$ specifies this expectation for any starting state $s$:\begin{equation}\label{eq:vpi}
    v_\pi(s) \doteq \CEpi{\sum_{j=1}^L \gamma^{j-1}R_j}{S_0\=s}, ~~~\forall s\in\S, 
\end{equation}
where the expectation is implicitly conditional on the actions being selected according to $\pi$. Value functions are defined for any $\pi$, but most commonly $\pi$ is the policy currently being followed by the agent and which gradually approaches an optimal policy (which maximizes \eqrf{vpi}).

A general value function (GVF; Sutton et al., 2011; Modayil, White \& Sutton, 2014) extends the idea of a value function in three ways. First, the reward $R_t$ is generalized to an arbitrary quantity to be added up, a \emph{cumulant} $C_t\d= c(S_t)$ for any \emph{cumulant function} $c:\S\ra\Re$. Second, the accumulation can stop not just at termination, but at a time with probability $\beta(S_t)$ for an arbitrary \emph{stopping function} $\beta: \S\ra[0,1]$. Third, at the time of stopping, $K$, there is a further \emph{stopping value} $z(S_K)$ added to the accumulation, for any \emph{stopping-value function} $z: \S\ra\Re$. Stopping is different from termination, as termination resets the state, whereas stopping has no effect on the state trajectory. At termination, the accumulation always stops ($\beta(\bot)\d= 1$) with zero stopping value ($z(\bot)\d= 0$). Formally, a GVF $v^{c,z}_{\pi,\beta}:\S\ra\Re$ is defined by
\begin{equation}\label{eq:v}
 v^{c,z}_{\pi,\beta}(s) \doteq \CEE{\pi,\beta\!}{\sum_{j=1}^{K} \gamma^{j-1} c(S_{j}) +\gamma^{K-1}z(S_K)\!}{\!S_0\=s}, ~~~\forall s\in\S,
\end{equation}
where the expectation is conditional on the actions being determined by $\pi$ and the stopping time $K\le L$ being determined by $\beta$.
The GVF gives the expected sum of the cumulant plus the stopping value if the policy were followed from $s$, stopping according to $\beta$.

To use GVFs to formulate subtasks, the superscript functions $c,z$ are taken as fixed and as defining the subtask, and the subscript functions $\pi,\beta$ are varied and taken as a possible solution. The subscript functions specify a policy and a stopping function, in other words, an \emph{option} (Sutton, Precup \& Singh, 1999).\footnote{Options as originally formulated also specified a set of states in which they could be initiated. We don't use that in this work so we elide it for simplicity.} Options are possible ways of behaving and stopping. If option $\pi, \beta$ were initiated in state $S_t$, then $A_t$ and subsequent actions would be selected according to $\pi$ until the option ended, or \textit{stopped}, according to $\beta$ at step $K$. To solve the subtask is to find an option which maximizes \eqrf{v}. 

The main task is a special case of a GVF task in which $C_{t}\d= R_{t}$ and stopping is not allowed ($\beta\d= 0$ or $z\d=-\infty)$. Shortest path subtasks are defined by $C_{t}\d= -1$ and $z(s)\d= 0$ at subgoal states and $\beta\d= 0$ or $z(s)\d= -\infty$ otherwise. GVF tasks includes all the common subtasks in the literature including those based on curiosity and intrinsic motivations (e.g., Baranes \& Oudeyer, 2013; Eysenbach et al., 2019).

\textit{Reward-respecting subtasks} are GVF tasks whose cumulant is identical to the reward and whose stopping values take into account the estimated value of the state stopped in. 
That is, for a reward respecting subtask, $c(S_t)\doteq R_t$, and $z$ is based on, but not identical to, $v_\pi$. 
The stopping values should not equal the estimated values because then the subtask would approximate the main task and solving it would probably add nothing new. Moreover, if there are multiple subtasks, each should have its own stopping values.

In this paper we focus on reward-respecting subtasks whose stopping values are designed to encourage solutions (options) that stop when a particular state feature is high.
Such \emph{subtasks of feature attainment} are appealling in several ways, but are not essential to this paper's main conclusions.
Other kinds of reward-respecting subtasks could have been used without otherwise affecting the STOMP progression.

To describe subtasks of feature attainment precisely, we need to describe our value-function approximation.
We assume that the state is represented as a feature vector $\x_t\d=\x(S_t)$ for some \emph{feature function} $\x:\S\ra\Re^d$ (with $\x(\bot)\d=\vec0$) which might be provided by a domain expert or might be hidden-unit activities of a neural network. We further assume that the approximation $\hat v$ of $v_\pi$ is linear in $\x(s)$ and a modifiable weight vector $\w\in\Re^d$:
\begin{equation}\label{eq:vhat}
  v_\pi(s) ~\approx~ \v(\x(s),\w) ~\doteq~ \w\tr\x(s) ~\doteq~ \sum_{i=1}^d w_i x_i(s), ~~~\forall s\in\S, 
\end{equation}
where $w_i$ and $x_i(s)$ are individual components of $\w$ and $\x(s)$, respectively.

Informally, to solve the reward-respecting subtask for attaining the $i$th feature is to find an option that obtains a lot of reward and stops when $x_i$ is high. More formally, let $i$ be a feature whose weight $w_i$ in the linear approximate value function \eqrf{vhat} varies over time, and let $\bw i$ denote one of its largest values, called the \emph{bonus weight}. The stopping-value function for the $i$th subtask is then defined as the estimated value, except using the optimistic bonus weight $\bw i$ in place of the usual weight $w_i$. That is, the $i$th subtask's stopping-value function is 
\begin{equation}
z^i(s) \doteq \w\tr\x(s) - w_i x_i(s) + \bw i x_i(s). \label{eq:z}
\end{equation}
The quantity $(\bw i -w_i) x_i(s)$ is sometimes called the \emph{stopping bonus} because it is the bonus for stopping in state $s$ beyond its (estimated) value on the main task. The stopping bonus is zero if $x_i(s)$ is zero; to get a large bonus, the option must stop in states of high estimated value in which feature $i$ is also high.
Note that $z^i$ does not actually depend on $w_i$; the inner product contains one term of $w_i x_i$ which cancels with the equation's second term, leaving $w_i$ effectively \emph{replaced} by the bonus weight. 
Also note we are using the index $i$ both as a subtask number when in the superscript position and as a feature number when in the subscript position. 
We use this convention throughout the paper.
Subtasks of this form are termed \textit{reward-respecting subtasks of feature attainment}.
 
Generally, it is only useful to construct subtasks for attaining a feature $i$ if $w_i$, its estimated contribution to value on the main task, is sometimes high and sometimes low. If $w_i$ never varied, then its static value could be learned once and never have to be changed by planning. As the ultimate use of all subtasks in the STOMP progression is for planning, such a subtask would never be useful.
If $w_i$ does vary, then its bonus weight is set to one of its higher values so that an option can be learned in preparation for the times at which it is high.


The reward-respecting subtask used in the illustrative example (Figure~\ref{fig:exp1}) was for the tabular feature for the hallway state, with bonus weight $\bw h \d= 1$, where $h$ denotes the index of the feature for the hallway state. The shortest-path subtask used $C_t\d=-1$ and stopped upon reaching the hallway or terminal goal states.

\section{Option learning}\label{sec:option_learning_feat_attainment}

In this section we specify the off-policy learning algorithms we use to approximate the optimal value functions and optimal options, which is the second step in the STOMP progression.


We describe these algorithms in a somewhat unusual way that lets us cover all the cases very compactly and uniformly, including the model learning cases in the next section. First we define a general TD (Temporal Difference) error function $\delta:\Re^4\times[0,1]\ra\Re$:
\begin{equation}\label{eq:delta}
    \delta(c, z, v, v', \beta) \doteq c + \beta z + \gamma(1\!-\!\beta)v' - v.
\end{equation}
Second, we define a general update procedure for learning with traces, which we call \textit{UpdateWeights\&Traces} (\UWT):

\def\gets{\leftarrow}

\begin{tabbing}
~~~~~~\=~~~~~~~\=~~~~\=\kill
~~~~~~\textbf{Procedure} \UWT\!\!$\bigl(\w, \e, {\nabla}, \alpha\delta, \rho, \gamma\lambda(1\!-\!\beta)\bigr)$:\+\+\\
  $\e \gets \rho (\e + {\nabla})$\\
  $\w \gets \w + \alpha\delta\e$\\
  $\e \gets \gamma\lambda(1\!-\!\beta)\e$
\end{tabbing}

\noindent
The first two arguments to \UWT are a weight vector and an eligibility-trace vector. These arguments are both inputs and outputs; the same pair are expected to be provided together on every time step. The weight vector is the ultimate result of learning. The eligibility trace is a short-term memory that helps with credit assignment. The third argument is usually a gradient vector with respect to the weight vector.
The fourth and sixth arguments to \UWT are scalars---the names of the formal arguments are just suggestive of their use.
Finally, the fifth argument is a scalar importance-sampling ratio used in off-policy learning (for on-policy learning it should be one).
Notice we do not use \UWT in a prescriptive way; nothing prevents us from using more efficient learning algorithms in each step of the STOMP progression. We use \UWT here to compactly describe the learning algorithms used in each step of the STOMP progression while also highlighting how similar they are.

\def\nablaw{\nabla_{\!\w}}
\def\nablatheta{\nabla_{\!\h}}
\def\nablathetai{\nabla_{\!\h^i}}

As an example, here is how these tools would be used to implement on-policy linear TD\la (Sutton, 1988) on the main task. 
First we would zero-initialize $\w$ and the corresponding eligibility-trace vector $\e\in\Re^d$. Then we would have the agent behave according to some policy $\pi$ and, on each time step in which $S_t$ is nonterminal, we would do:
\begin{align}
&\delta \gets \delta(R_{t+1},0, \v(\x_t,\w), \v(\x_{t+1},\w), 0), \text{~and} \label{eq:ex_td}\\
&\UWT(\w, \e, \nablaw\v(\x_t,\w), \alpha\delta, 1, \gamma\lambda), \label{eq:critic-ex}   
\end{align}
where $\alpha$ and $\lambda$ are step-size and bootstrapping parameters respectively.
Note that in the linear case, $\nabla\v(\x_t,\w)$ in \eqrf{critic-ex} is just $\x_t$.

As another example, suppose the policy $\pi$ is parameterized by $\h\in\Re^{d'}$ and we want to learn it as well, in an actor-critic algorithm (Sutton, 1984; Sutton et al., 2000; Konda \& Tsitsiklis, 2003). This would be achieved by invoking \UWT one more time on each step, immediately after \eqrf{critic-ex}:
\begin{equation}\label{eq:actor-ex}
    \UWT(\h, \e', \nablatheta\ln\pi(A_t|S_t,\h), \alpha'\delta, 1, \gamma\lambda'),
\end{equation}
where $\e'\in\Re^{d'}$ is another zero-initialized eligibility-trace vector, and $\alpha'$ and $\lambda'$ are step-size and bootstrapping parameters for the actor.

Now we show how to use these tools in the second step of the STOMP progression to learn the value functions and options for the subtasks from off-policy experience.
Let $\T\subset\{1, \ldots, d\}$ be the set of features for which we have subtasks. Each subtask $i\in\T$ will have a value-function weight vector $\w^i\in\Re^d$ and a policy weight vector $\h^i\in\Re^{d'}$ such that $\v(\x(s),\w^i)\d=\w^i\tr\x(s)\approx v^{r,z^i}_{\pi^i,\beta^i}(s)$, as in \eqrf{v}, where $r(S_t)\d=R_t$, $\pi^i\d=\pi(\cdot|\cdot,\h^i)$, and 
\begin{equation}\label{eq:stopping}
\beta^i(s) \doteq
\begin{cases}
    1,& \text{if } z^i(s) \geq \w^i\tr\x(s);\\
    0,              & \text{otherwise,}
\end{cases}~~~~\forall i\in\T, s\in\S, 
\end{equation}
with $\beta^i(\bot)\d=1$. 
Under this definition, the option $\pi^i,\beta^i$ stops in state $s$ if the stopping value, $z^i(s)$, which is the estimated main-task value using the bonus weight $\bw i$ instead of $\w_i$, is greater than or equal to the estimated subtask value.
That is, the option does not stop if the estimated subtask value of continuing is better than the stopping value.

We are learning off-policy, so we need to use the importance sampling ratios
$\rho^i_t \doteq \frac{\pi^i(A_t|S_t)}{\mu(A_t|S_t)}$, 
where $\mu:\A\times\S\ra [0,1]$ is the \textit{behavior policy} (the policy actually used to select actions, as opposed to the policies being learned about).
For each subtask $i\in\T$, in addition to $\w^i$ and $\h^i$ we will also have eligibility-trace vectors $\e^i\in\Re^d$ and ${\e'}^i\in\Re^{d'}$, all initialized to zero. Then, on each time step on which $S_t$ is nonterminal, for each $i\in\T$ we do:
\begin{align}\label{eq:actor-critic-learning}
&\delta \gets \delta\bigl(R_{t+1},z^i(S_{t+1}), \v(\x_t,\w^i), \v(\x_{t+1},\w^i), \beta^i(S_{t+1})\bigr), \nonumber\\
&\UWT\bigl(\w^i, \e^i, \nablaw\v(\x_t,\w^i), \alpha\delta, \rho^i_t, \gamma\lambda\!\left(1\!-\!\beta^i(S_{t+1})\right)\bigr), \text{~and}   \\
&\UWT\bigl(\h^i, {\e'}^i, \nablathetai\ln\pi(A_t|S_t,\h^i), \alpha'\delta, \rho^i_t, \gamma\lambda'\!\left(1\!-\!\beta^i(S_{t+1})\right)\bigr). \nonumber
\end{align}

\noindent 
Under this algorithm, the learned approximate values $\v(\x(s),\w^i)\d=\w^i\tr\x(s)$ come to approximate the \textit{optimal subtask values} $v_*^i(s) \doteq \max_{\pi,\beta} v_{\pi,\beta}^{r,z^i}(s)$, for all $s\in\S$ and $i\in\T$, 
and the options $\langle\pi(\cdot|\cdot,\h^i), \beta^i\rangle$ come to approximate corresponding optimal options.
In the tabular case, the approximations become exact with sufficient exploration and if the step sizes are decreased~appropriately.

We applied this algorithm to the hallway-feature-attaining subtask introduced as an illustrative example in Figure~\ref{fig:exp1}, using a behavior policy that selected all four actions with equal probability: $\mu(a|s)\doteq 0.25$, for all $s\in\S, a\in\A$.
The state-feature vectors were one-hot, with $d\=72$ for the 72 non-terminal grid cells. 
The policy was of the softmax form with linear preferences: 
\begin{equation}\label{eq:softmax}
\pi(a|s,\h)\doteq\frac{e^{\h\tr\xa(s,a)}}{{\textstyle \sum_{b}}e^{\h\tr\xa(s,b)}},  
\end{equation}
where the state-action feature vectors $\xa(s,a)\in\Re^{d'}$ were again one-hot ($d'\=288$).
The weight vectors were all initialized to zero, so the initial approximate value function was everywhere zero and the initial policy was equi-probable random.
The parameters where $\alpha\=\alpha'\=0.1$ and $\lambda\=\lambda'\=0$.

\begin{figure}
\centering
\vspace{-.4in}
\includegraphics[width=0.9\textwidth]{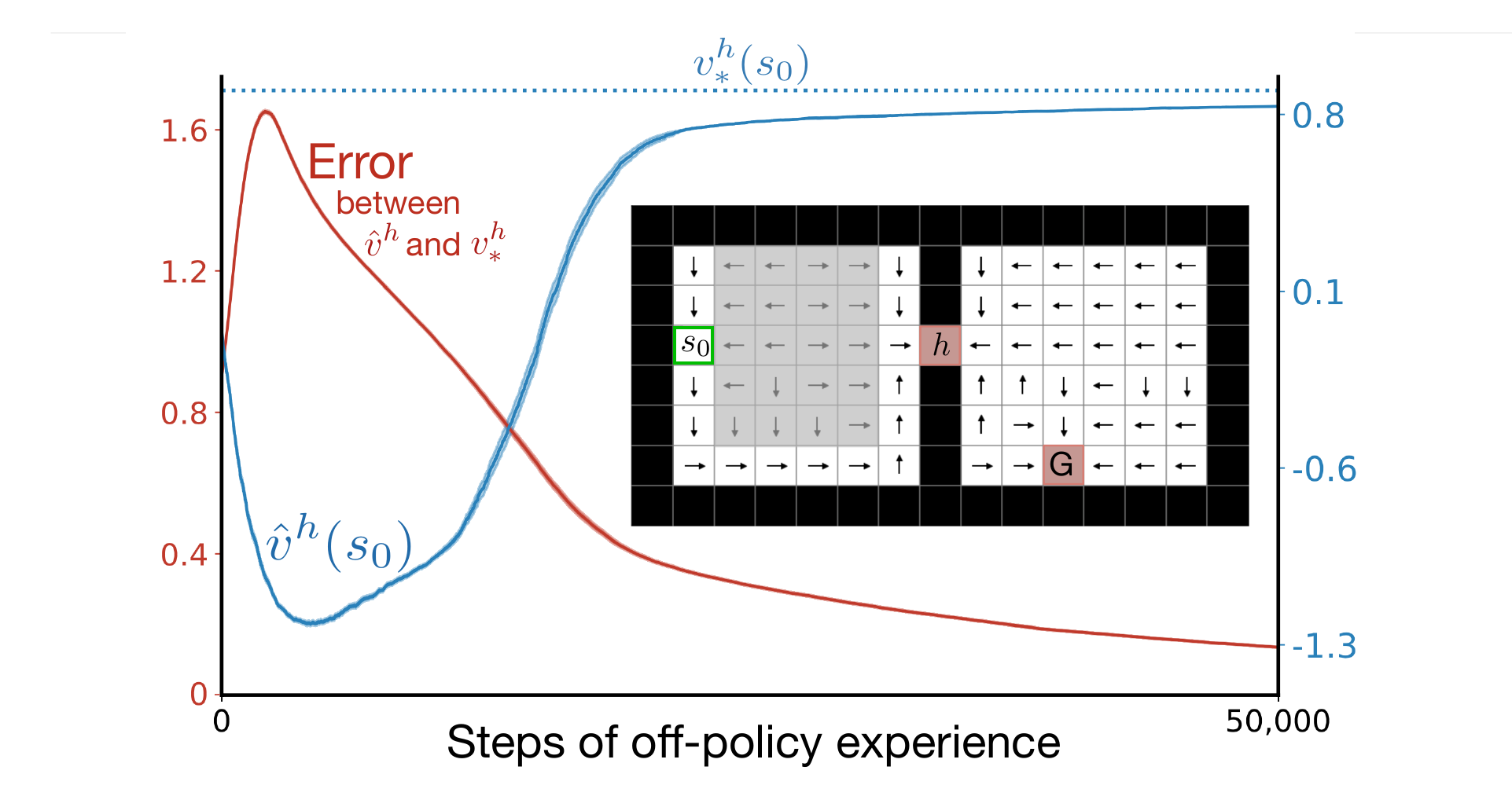}
\vspace{-.05in}
\caption{\textbf{Option learning experiment}. Algorithm \eqrf{actor-critic-learning} finds an optimal option and its value function for the reward-respecting subtask for attaining the hallway feature.
The red line shows the root-mean-squared error between the estimated and optimal values decreasing toward zero (left scale).
The blue line shows the estimated value on the hallway subtask approaching its optimal value of $v^h_*(s_0)=\gamma^{11}\approx .895$ (right scale).
These data are averages over 100 runs and the shading is one standard error. 
\textbf{Inset}, the arrows show a learned option policy and the red cells show the states in which that option stops.}
\vspace{-.0in}
\label{fig_option_policy}
\end{figure}

As a measure of the quality of the learned value function, we recorded the root mean squared error (RMSE) between the estimated and optimal state values at each step, averaged over all states, and as a measure of the quality of the policy, we recorded the estimated value of the start state $\v^h(s_0)\d=\v(s_0,\w^h)$ at each step (see Figure~\ref{fig_option_policy}). The error from optimal values first rises as $\v^h$ approximates the value of the random policy, then falls toward zero as the option becomes optimal for the hallway-attaining subtask. The estimated value $\v^h(s_0)$ starts at zero, then falls as the near-random policy wanders into the gray field of negative reward, and finally rises toward $v^h_*(s_0)\=\gamma^{11}\approx .895$, the value of the start state under the optimal policy for the hallway subtask.

These results show that the actor-critic algorithm \eqrf{actor-critic-learning} is able to  learn the correct option policy and value function from off-policy data. Notice that the option stops either at the hallway state or the goal state (shown in red in Figure~\ref{fig_option_policy}). Reward-respecting options tradeoff rewards, the value of the state, and the stopping bonus.  In the right room, close to the hallway state it is more beneficial to go to the hallway state because the stopping bonus plus the value of that state is greater than the value of directly going to the goal state. In states closer to the goal state, it is better to directly go to the goal state rather than back to the hallway. 

Note that the policy and value function for the main task can also be learned by the actor-critic algorithm \eqrf{actor-critic-learning} simply by considering the main task to be one more subtask, say subtask 0, with $\beta^0\d=0$.

\section{Model learning}\label{sec:model_learning}

In this section we describe the third step in the STOMP progression: learning a model of the environment's action and option transitions. Recall that an option is a pair, $o\doteq(\pi_o, \beta_o)$, consisting of a policy and a stopping function. Actions are a special case of options in which the policy $\pi_o$ always selects the action and the stopping function always stops, $\beta_o(s)\=1$, for all $s\in\S$.
Let $\O(s)$ denote the set of options (including actions) available in state $s$, and let $\O$ denote the set of all options (unioning over all states). 
With a slight abuse of notation, we allow these sets of options to include the indices of state-features $i\in\T$ for which the agent has learned an option; when such a feature index appears in a position that is expecting an option, we mean the option corresponding to that feature.

The \textit{ideal} model is expressed in terms of the underlying environment states $s\in\S$ and exactly matches the true underlying dynamics. Like all models, it is comprised of a reward part and a state-transition part.
The reward part is a function $r: \S\times \O\rightarrow\Re$ returning the expected cumulative discounted reward if the option were executed starting from the state:
\begin{equation}
 r(s,o) \doteq \CEE{\pi_o,\beta_o}{{\sum_{t=1}^K} \gamma^{t-1} R_t}{S_0\=s}, 
 ~~~\forall s\in\S, o\in\O(s), 
 \end{equation}
where the expectation is conditional on actions being selected by $\pi_o$ and the stopping time $K$ being determined by $\beta_o$. 
The state-transition part of an ideal model is a function ${p}: \S\times\S\times\O\rightarrow[0, 1]$ returning, for each state $s$ in which an option might be started, the probability of stopping in each state $s'$, discounted by the time until stopping:
\begin{equation}\label{eq:p}
 p(s'|s,o) \doteq \sum_{t=1}^\infty \gamma^{k}\CP{K\=t,S_{t}\=s'\!}{\!S_0\=s}, 
~~~\forall s\in\S, o\in\O(s), 
\end{equation}
where the probability is conditional on the actions being selected according to the policy of option $o$ and $K$ being determined by its stopping function. Note that we write the $p$ function with a $|$, suggesting that it is a probability distribution, but for $\gamma<1$ it is not. This precise form is dictated by the requirement that option models and action models be interchangeable in planning methods such as value iteration (see next section).

Approximate models do not in general have access to environmental states, but instead must work with a representation of state constructed by the agent, which might be called the \textit{agent state}. Unlike the environment state, the agent state is generally not a Markov summary of the past. In this paper we assume the agent state is a feature vector $\x\in\Re^d$ which can be determined from the environment state by a known function $\x: \S\ra\Re^d$.
The reward part of an approximate model of an option is a function $\hat r: \Re^d\times\O\ra\Re$ such that
\begin{equation}\label{eq:rhat}
    \r(\x(s),o) \approx r(s, o), ~~~~\forall s\in\S, 
\end{equation}
weighted over some distribution of states such as the state distribution under the behavior policy.
The state-transition part of an approximate model can take several forms (Kudashkina, 2022).
In this paper we use an \textit{expectation} model (Wan et al., 2019), in which the state-transition part is a function $\n:\Re^d\times\O\ra\Re^d$ such that
\begin{align}\label{eq:nhat}
    \n(\x(s),o) &\approx \sum_{s'\in\S} p(s'|s,o) \x(s')
    =\CEE{o}{\gamma^K\x(S_{t+K})}{S_t\=s}, 
 ~~~\forall s\in\S,  
\end{align}
under some state weighting, perhaps given by the behavior policy, and where the expectation is conditional on the actions being selected according to $o$'s policy and the stopping time $K$ being determined by $o$'s stopping function.

For the illustrative example in Figure 1, we learned models of the four options corresponding to  actions and of the one reward-respecting option for attaining the hallway state.
The approximate model was linear in the state-feature vector, meaning that
\begin{equation}
    \r(\x,o) \doteq \w_r^o\tr\x ~~~~~~\text{and}~~~~~~
    \n(\x,o) \doteq \W^o\x, 
\end{equation}
where each $\w_r^o\in\Re^d$ is a learned weight vector, 
and each $\W^o$ is a $d\times d$ matrix, with rows $\w^o_j$.
In this tabular problem the states were represented by one-hot feature vectors ($d=|\S|=72$).
To learn the weights, each weight vector (the $\w^o_r$ and the $\w^o_j$, for $o\in\O, j\=1, \ldots, d$) was paired with an eligibility trace vector, $\e^o_r$ or $\e^o_j$. All these vectors were initialized to zero.
The agent wandered throughout the two rooms following the equi-probable random policy for 50,000 time steps.
On each transition ($S_t, A_t, R_{t+1}, S_{t+1}$) for which $S_t$ was non-terminal, for each $o\in\O$, we did:
\begin{align}\label{eq:model-learning}
&\delta \gets \delta\bigl(R_{t+1},0, \r(\x_t,o), \r(\x_{t+1},o), \beta^o(S_{t+1})\bigr) \nonumber\\
&\UWT\bigl(\w_r^o, \e_r^o, \nabla\r(\x_t,o), \alpha_r\delta, \rho^o_t, \gamma\lambda\bigl(1\!-\!\beta^o(S_{t+1})\bigr)\bigr) \nonumber\\ 
&\text{and, for each $j=1, \ldots, d$:}   \\
&~~~~~~\delta \gets \delta\bigl(0, x_{j,t+1}, \hat n_{j}(\x_t,o), \hat n_{j}(\x_{t+1},o), \beta^o(S_{t+1})\bigr)  \nonumber\\
&~~~~~~\UWT\bigl(\w_{j}^o, \e^o_{j}, \nabla\hat n_{j}(\x_t,o), \alpha_p\delta, \rho^o_t, \gamma\lambda\bigl(1\!-\!\beta^o(S_{t+1})\bigr)\bigr), \nonumber
\end{align}
where $x_{j,t+1}$ denotes the $j$th component of $\x_{t+1}$ and $\hat n_j$ denotes the $j$th component of the vector returned by $\n$.
This way of using TD\la algorithms to learn a temporally abstract model of the world was introduced by Sutton (1995; see also Sutton and Barto, 2018, Section 17.2)
The parameters were $\alpha_r\=\alpha_p\=0.1$ and $\lambda\=0$.

\begin{figure}
\centering
\vspace{-.05in}
\includegraphics[width=\textwidth]{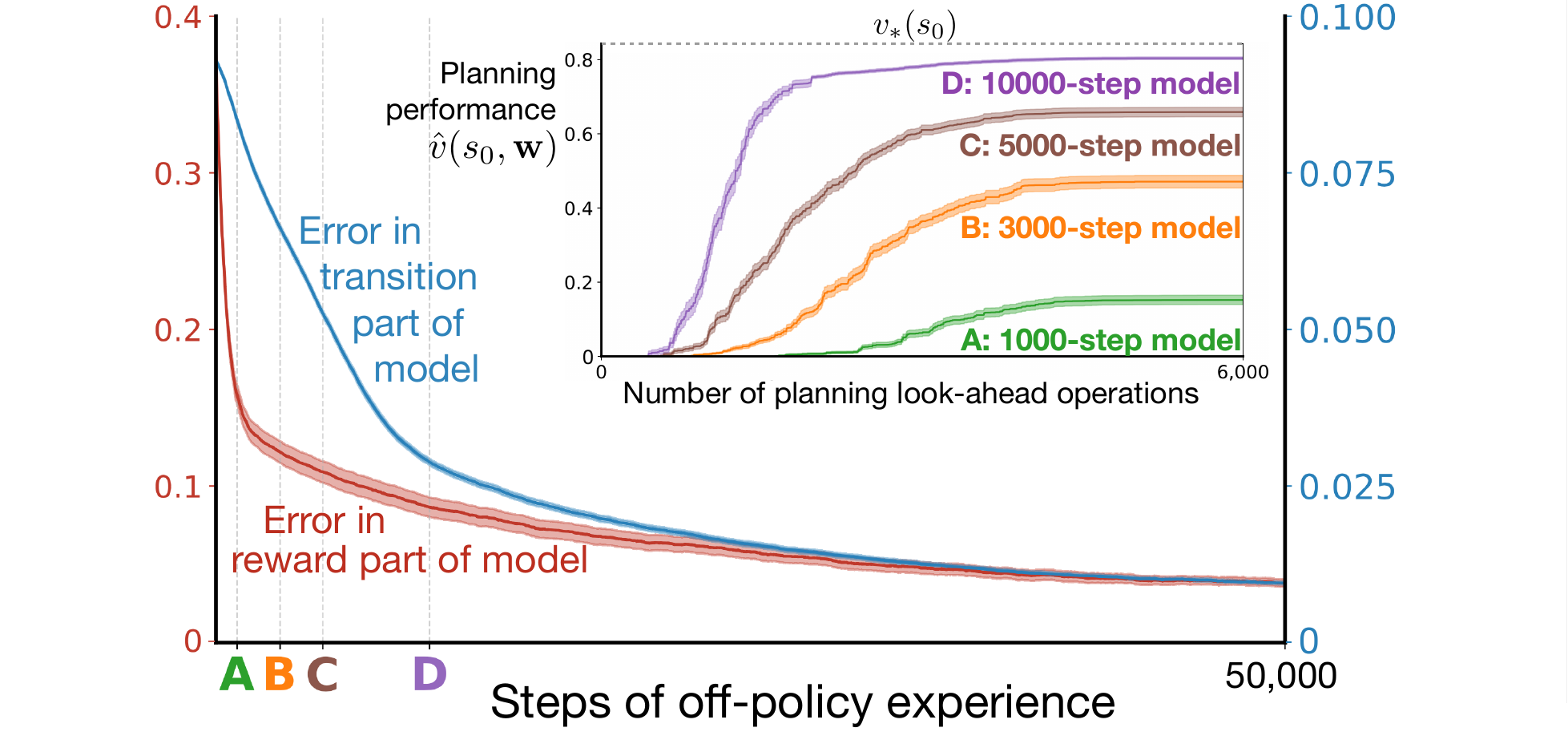}
\caption{\textbf{Model learning experiment}. The root-mean-square error in the model of the hallway-attaining option falls toward zero under off-policy training. \textbf{Inset}: The models at various times A-D are assessed for their utility in planning; the later, more-accurate models enable faster planning to better policies. All lines are averages over 100 runs.}
\label{fig_model_learning_exp2}
\end{figure}

The procedure described above allows us to efficiently learn both transition and reward models, as shown in Figure~\ref{fig_model_learning_exp2}. We also evaluated the impact of planning with imperfect models, using the planning algorithm described in the next section. The results are shown in the inset plot. Planning leads to near-optimal performance after the model has been trained for about 20,000 steps. Further training improves model error but does not significantly improve planning in this environment. The final model, after 50,000 steps, was used for the planning results in Figure 1. 

A linear expectation model is not the most general form of an environmental model, but still may be a good choice.
More general would be a world model in which the transition-part is
able to produce result states with the correct joint distribution. These {\em distribution} models are important theoretically and have been
used effectively when the possible distributions can be assumed to be of a special form (e.g., Gaussian), as in the PILCO method~(Deisenroth \& Rasmussen, 2011). Generally, distribution models are very large objects and would be hopelessly unwieldy for large $d$ (unless the distributions are of particular forms, such as Gaussians). Fortunately, if the value function is linear in the state features, then there is no loss of generality when an expectation model is used in planning (see Wan et al., 2019, Kudashkina, 2022). 

The deeper issue is that no model can be complete and accurate, as the world is much larger and more complex than the agent (Javed \& Sutton, in preparation). An expectation model is one strategy for accepting this gracefully.

\section{Planning with options}\label{sec:planning_with_options}

Our planning method approximates \emph{asynchronous value iteration}, a classical operations-research planning algorithm for finite MDPs, extended to options (Sutton, Precup \& Singh, 1999). 
In the tabular (non-approximate) algorithm, the state-value estimates $V(s)$, for all $s\in\mathscr{S}$, are initialized arbitrarily and then updated one-by-one, in some sequence, by:
\begin{equation}\label{eq:VI}
 V(s) \leftarrow \max_{o\in\mathscr{O}(s)} \left[ r(s,o) + \sum_{s'\in\S} p(s'|s,o)V(s') \right], 
\end{equation}
where $r$ and $p$ are the reward and state-transition parts of the ideal model of option $o$ (as defined in the previous section), and $\mathscr{O}(s)$ is the set of options considered in state~$s$, which may include all or some of the primitive actions available. If $\mathscr{O}(s)$ is exactly the primitive actions, then $\hat p(s'|s,o)$ is exactly the state-transition probabilities, times~$\gamma$, and the general form \eqrf{VI} reduces to classical value iteration.

In our planning method, \emph{approximate value iteration}, the estimated value function is maintained not as a table $V(s)$, but as a parameterized form $\hat v(\x(s), \w)$ with feature function $\x:\S\ra\Re^d$ and weight vector $\w\in\Re^d$ with $d\ll|\mathscr{S}|$. 
Moreover, the planner will iterate over state-feature vectors $\x\in\Re^d$ instead of environmental states $s\in\S$.
In this paper we use a linear form $\hat v(\x,\w)\doteq \w\tr\x$, which combines favorably with expectation models, but in general any differentiable parameterized form could be used.
The weight vector is initialized arbitrarily and then updated, for each state-feature vector $\x$ in some sequence of state-feature vectors, by
\begin{equation}\label{eq:AVI}
\w \leftarrow \w +\alpha \Bigl[\max_{o\in\mathscr{O}(\x)} \left[\r(\x,o) + \hat v(\n(\x, o), \w)\right] - \hat v(\x,\w)\Bigr]\nabla_{\!\w} \hat v(\x,\w),
\end{equation}
where $\alpha>0$ is a step-size parameter, and $\hat r$ and $\n$ are the reward and transition parts of an approximate model as described in the previous section. 

The quantity $\r(\x,o) + \hat v(\n(\x, o), \w)$ in \eqrf{AVI} is the called the \textit{backed-up value} of the state represented by $\x$, when projected ahead by the model of option $o$, using the approximate value function given by the weights $\w$. Computing the backed-up value for a state and option counts as one \emph{planning look-ahead operation} in our result figures (the x-axis of, e.g., Figure 1). The backed-up value is a target, analogous to the quantity in brackets in \eqrf{VI}. Equation \eqrf{AVI} is a standard stochastic-gradient-descent update rule toward the backed-up value (ignoring the effect of the update on the target, as is commonly done in reinforcement learning). Note that the sum in \eqrf{VI} over $|\S|$ terms has been replaced in \eqrf{AVI} by a single call to $\n$ whose complexity is linear in the number of model parameters (e.g., only $d^2$ for $d\ll|\S|$ in the linear case). As discussed earlier, this can be done without introducing any additional approximation error if the value function is linear in the state-feature vector $\x$ (Wan et al., 2019). 

Approximate value iteration \eqrf{AVI} was applied to the two-room gridworld to obtain the results in Figure~\ref{fig:exp1}. We sampled states $s$ randomly from the full state set $\S$, computed their feature vectors $\x(s)$, and then performed \eqrf{AVI} on each state in sequence. The weight vector $\w$ was initialized to zero. The models of the actions and options were those learned after 50,000 random steps by off-policy methods as described in the preceding section. 
The step-size parameter was $\alpha\=1$.
Additional planning results with incompletely-learned models are shown inset in Figure~\ref{fig_model_learning_exp2}.

In this paper our presumption is that the model of the options will normally be accurate and stable, while the approximate value function will not have been previously fully learned and will not be stable (as otherwise no further planning would be necessary). These assumptions are noteworthy and deserve examination. Certainly there are cases where they are appropriate. In Chess, for example, the model of the game's dynamics is known completely, and could have been learned, but the value of almost all states can only be approximated, and the planning problem is so large that it is never completely solved and the state values are never all known exactly.

The primary definition of a useful option is one whose model takes the maximum in~\eqrf{VI} or \eqrf{AVI} at some state and thus makes a difference in planning. To make the backed-up values large (and thus more likely to take the maximum), we certainly seek options for which $r(s,o)$, the cumulative reward during the option, is large. This is the main point about reward-respecting options. We also want the second term of the backed-up value to be large, which is achieved if the option terminates in states of high approximate value. This is the reason for optimistic bonus weights. We seek options that produce high rewards and that drive the environment to states that, occasionally, have high value.

\section{Stopping bonuses matter}\label{sec:bonus-weights}

In Section 2 we defined reward-respecting subtasks for attaining any feature $i$ with respect to a \emph{bonus weight}, $\bw i$, that determines a \emph{stopping bonus}, $(\bw i - w_i)x_i(s)$, for stopping in state $s$ when feature $x_i(s)$ is high. The bonus weight controls the size of the stopping bonus and thereby the tradeoff between attaining feature $i$ and attaining reward, and thus the bonus weight is an important parameter of the subtask. 
So far we have considered only the case in which $\bw i \d= 1$. We now evaluate the impact of different choices of the bonus weight and, in particular, show that very large bonus weights result in subtasks whose solutions closely approximate shortest-path options.

\begin{figure}[b]
\vspace*{-.17in}
\centering
     \includegraphics[width=0.85\textwidth]{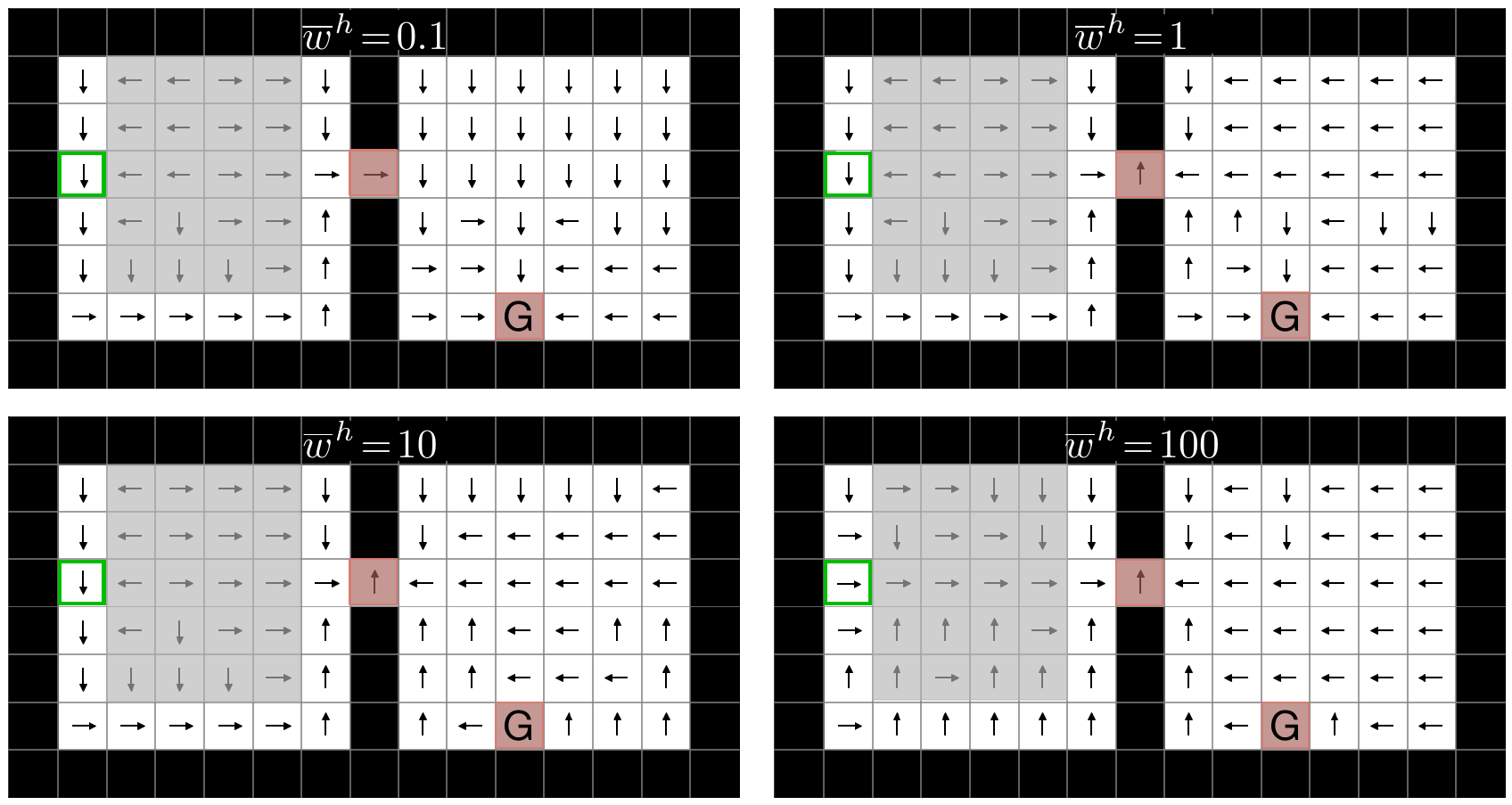}
     \caption{\textbf{Reward-respecting options} learned for different values of the bonus weight $\bw h$.}
     \label{fig:exp3_options_kappa}
\end{figure}

Figure~\ref{fig:exp3_options_kappa} shows the reward-respecting options learned for the two-room gridworld for various values of $\bw h \in \{0.1, 1, 10, 100\}$. The $\bw h\d=1$ case is the one we have already seen; its learned option takes a roundabout path that avoids the gray region where $-1$ rewards are received. For the largest bonus weight, $\bw h \d= 100$, on the other hand, the learned option takes shortest paths to the hallway state, while for $\bw h \d= 10$ the learned option exhibits intermediate behavior. Note that the policies are stochastic; the arrows in the figure show just the action with the \emph{largest probability} of being selected. The choice of $\bw h \d= 0.1$ is still optimistic for this task because at the time these options were learned the estimated value function for the main task was still at its initial value of zero. In this case, in the first room the learned option still avoids the region of negative reward, but in the second room it is more directed towards the goal state and less toward the hallway state than in the other three cases.
As expected, small values of $\bw i$ make attaining feature $i$ less important.

\begin{figure}
\vspace{-.05in}
    ~~~\includegraphics[width=.8\textwidth]{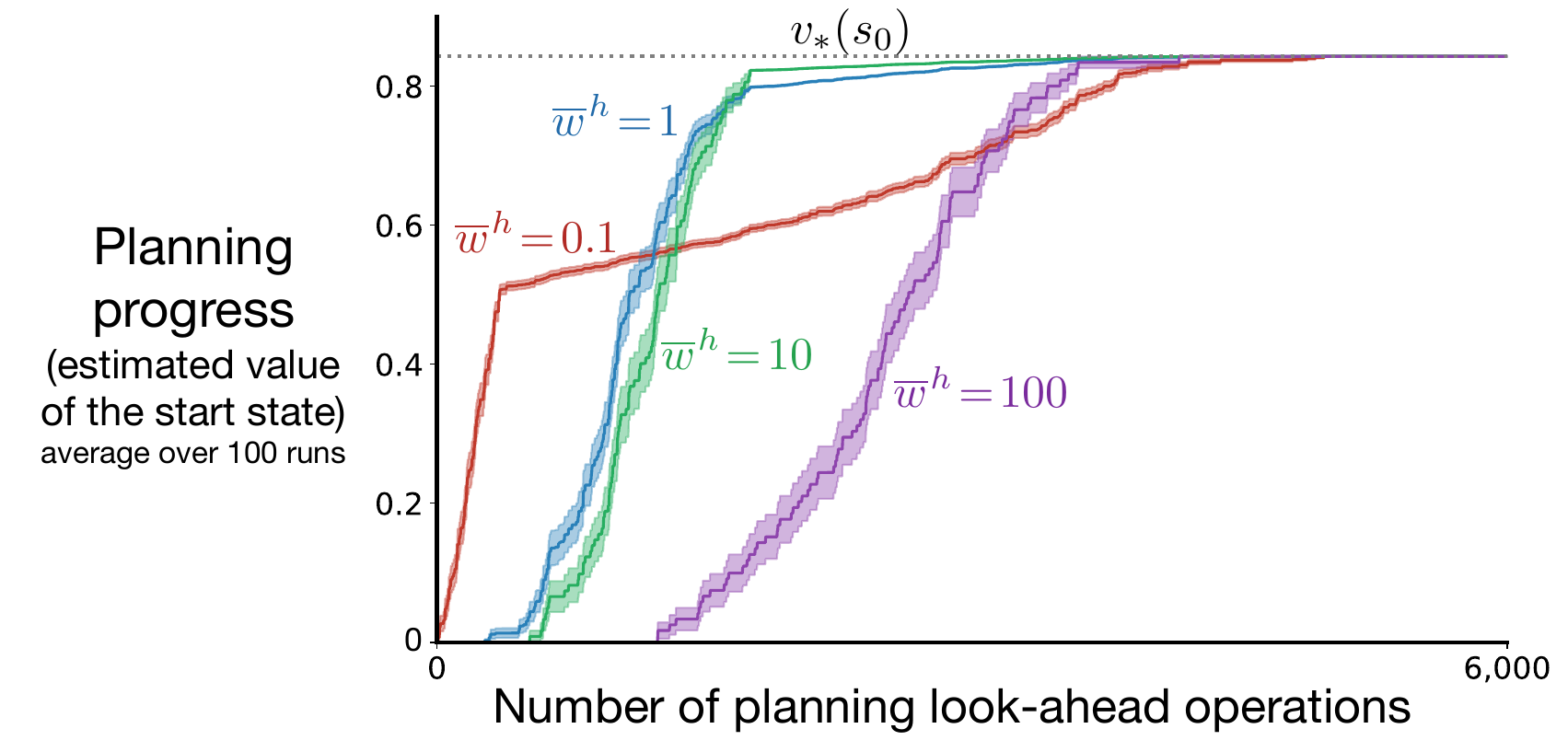}
    \caption{\textbf{Impact of the bonus weight} $\bw h$ on the efficiency of planning.}
    \label{fig:bonus-weight-plot}
\vspace{-.1in}
\end{figure}

For each of the four options, we continued with the other steps of the STOMP progression. We learned a model of the option, and then applied approximate value iteration to plan with the model. The progress of planning in the four cases was assessed, as before, by the estimated value of the start state (see Figure~\ref{fig:bonus-weight-plot}). The blue ($\bw h\d=1$) and purple ($\bw h\d=100$) lines replicate results from Figure 1; a reward-respecting subtask can result in much faster planning than a shortest-path subtask. The green line ($\bw h\d=10$) suggests that the advantage of reward-respecting subtasks is robust to the choice of the bonus weight. The $\bw h\d=0.1$ case is mixed; initially planning is fastest, presumably because the learned option itself closely approximates the optimal policy, but in the longer term planning is retarded according to this measure.

\section{STOMP in a larger, stochastic gridworld}~\label{sec:final_experiments}

In this section we further test the ideas of reward-respecting subtasks and the STOMP progression by applying them with multiple subtasks and options in a larger problem with stochastic dynamics. We also compare and contrast the reward-respecting approach with that of eigenoptions (Machado et al., 2017; 2018), the option-critic architecture (Bacon, Harb \& Precup, 2017), and, again, shortest-path options. For each of these four ways of producing options, we learn models of their options and use the models for planning exactly as described earlier in this paper. 

\begin{figure}[b]
  \centering
     \includegraphics[width=1.1\textwidth]{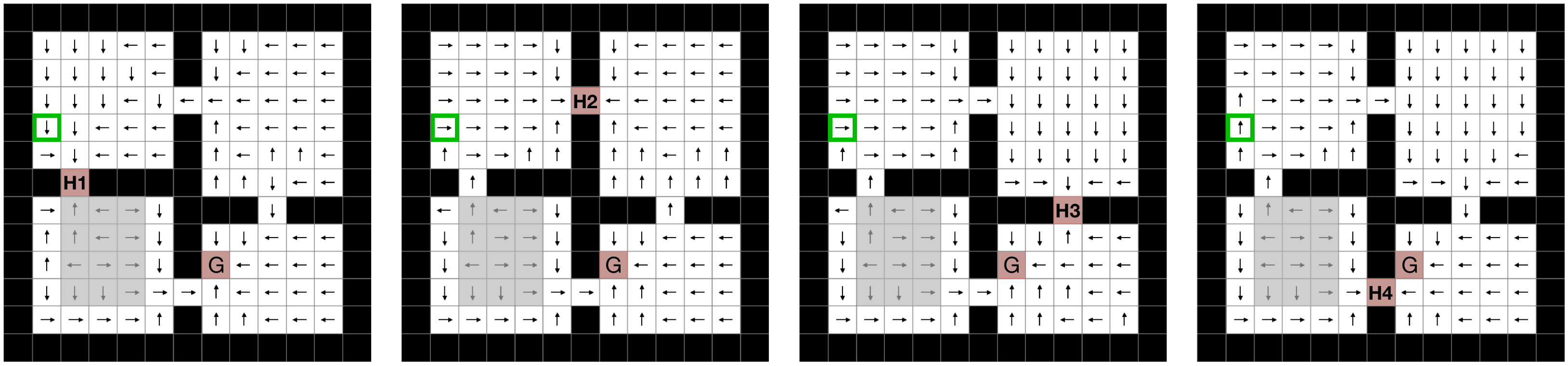}
     \caption{\textbf{The four-room gridworld and four reward-respecting options} learned for attaining its hallway states H1--H4. The start state is highlighted in green. The arrows show the action most favored by the learned option in each state (the actual policies are stochastic), and the red cells are states in which the option deterministically stopped.}
     \label{fig:fourrooms_options}
\end{figure}

The larger problem used in these comparisons is the four-room episodic gridworld depicted in each of the four parts of Figure~\ref{fig:fourrooms_options}, with a start state in the upper-left room (highlighted in green) and a terminal goal state in the lower-right room. As in the two-room gridworld, a reward of $+1$ is received on reaching the goal, which ends the episode, and passing through the gray region produces a reward of $-1$ per step, while all other transitions produce a reward of zero. Again the discount factor is $\gamma = 0.99$, and again there are four actions for moving in each one of the four directions, but now they are \emph{stochastic}, moving in the expected direction only with probability $\nicefrac{2}{3}$, and in one of the other three directions with probability $\nicefrac{1}{9}$. If the direction of movement (after stochasticity) is blocked by a wall, then the agent's location is unchanged.

\begin{figure*}[t]
     \centering
     \includegraphics[width=\textwidth]{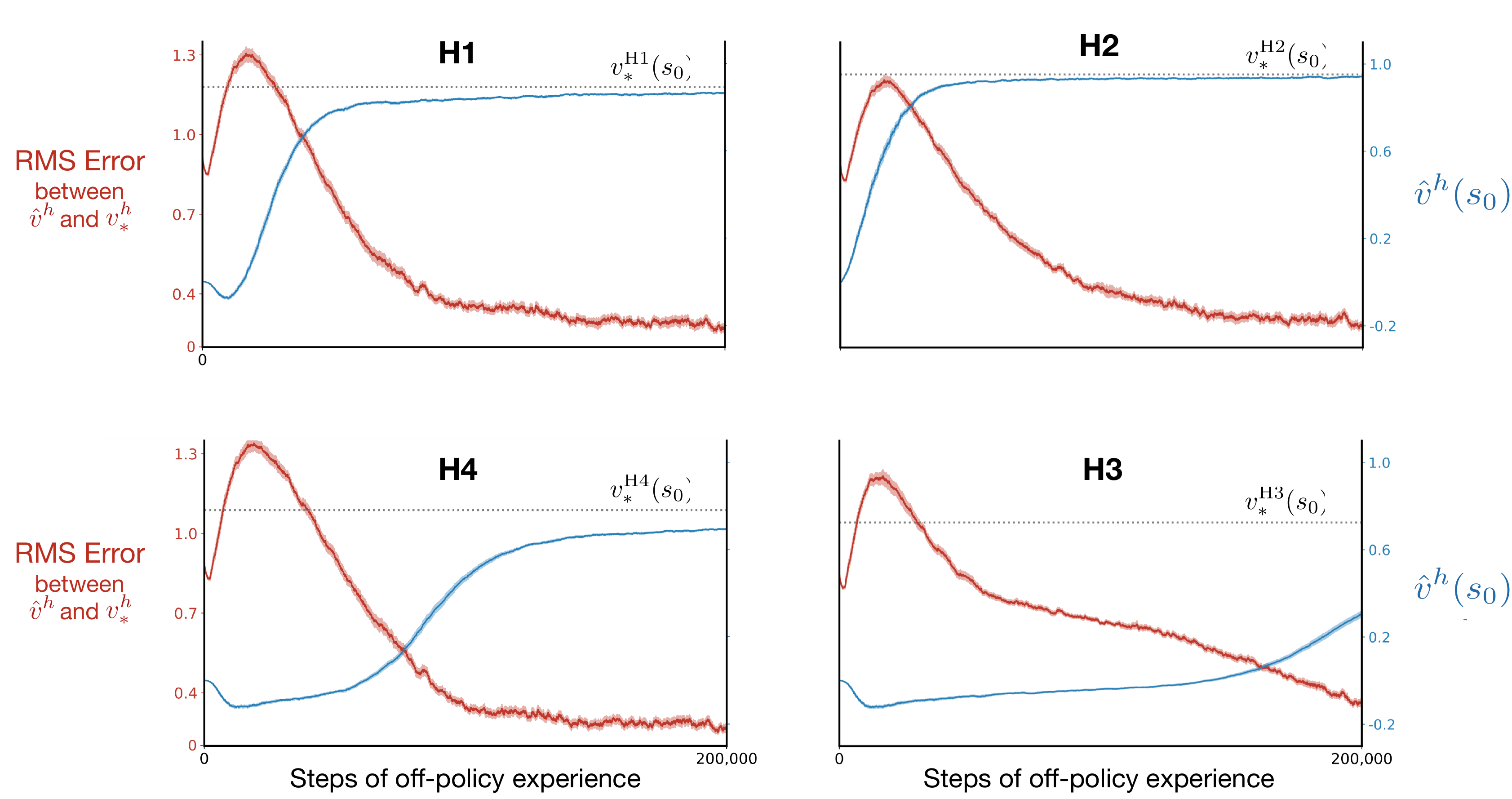}
     \caption{\textbf{Off-policy learning} for reward-respecting options in the four-room gridworld.}
     \label{fig:4option-learning}
\end{figure*}

For the reward-respecting approach, we defined four reward-respecting subtasks of feature attainment, each directed toward attaining the feature for one of the four hallways states H1--H4, and learned close approximations to their optimal options using the algorithms described in Section 3. The bonus weight for all four subtasks was $\bw i\=1$. 
The learned options are depicted in Figure~\ref{fig:fourrooms_options}---the policy by arrows and the stopping states by red cells. For the most part, the options took the shortest path to the hallway or the goal state, but the negative reward region caused some options to prefer a longer path. In particular, the options often took a roundabout way around the gray region in the lower-left room. The step-size parameter was $\alpha\=0.05$. Figure \ref{fig:4option-learning}
shows the progression of learning of the four options. 

\emph{Eigenoptions} (Machado et al., 2017; 2018) are defined as the solutions to subtasks with intrinsic rewards unrelated to the reward of the main task. The intrinsic rewards are constructed from the successor representation as an approximation to the graph Laplacian of the interconnection topology of the environment. We defined four subtasks corresponding to the first four eigenvectors (by largest eigenvalue) of the approximate graph Laplacian. In our notation, the $i$th subtask was to maximize the GVF \eqrf{v} with $z(S_K)\doteq 0$ and $C_t \doteq \e_i\tr\bigl(\x(S_t)-\x(S_{t-1})\bigr)$, where $\e_i$ is the $i$th eigenvector. Stopping is viewed as a special action (Machado et al., 2018) whose action value is defined to be zero (and thus need not be learned). In our experiments we learned the value of the other actions by one-step Q-learning with a random behaviour policy. Finally, the options were defined as greedy with respect to the action values.

The \emph{option-critic architecture} (Bacon, Harb \& Precup, 2017) uses an alternative way of discovering options without explicit subtasks or reward signals other than the main-task rewards. The options are initialized randomly and then climb the gradient of a global objective function. We used a re-implementation of the option-critic architecture that is part of DeepMind's software suite. For each run, that software produces a new set of options from experience solving the main task. 

Finally, four shortest-path options were approximated for the hallway states of the four-room gridworld in the same way as described earlier for the smaller gridworld. For each hallway, a GVF subtask was created with $C_t\d=-1$ for all $t$ and $z(s)\d=0$ at the hallway state and $-\infty$ otherwise.

\begin{figure}[b]
    \includegraphics[width=1.2\textwidth]{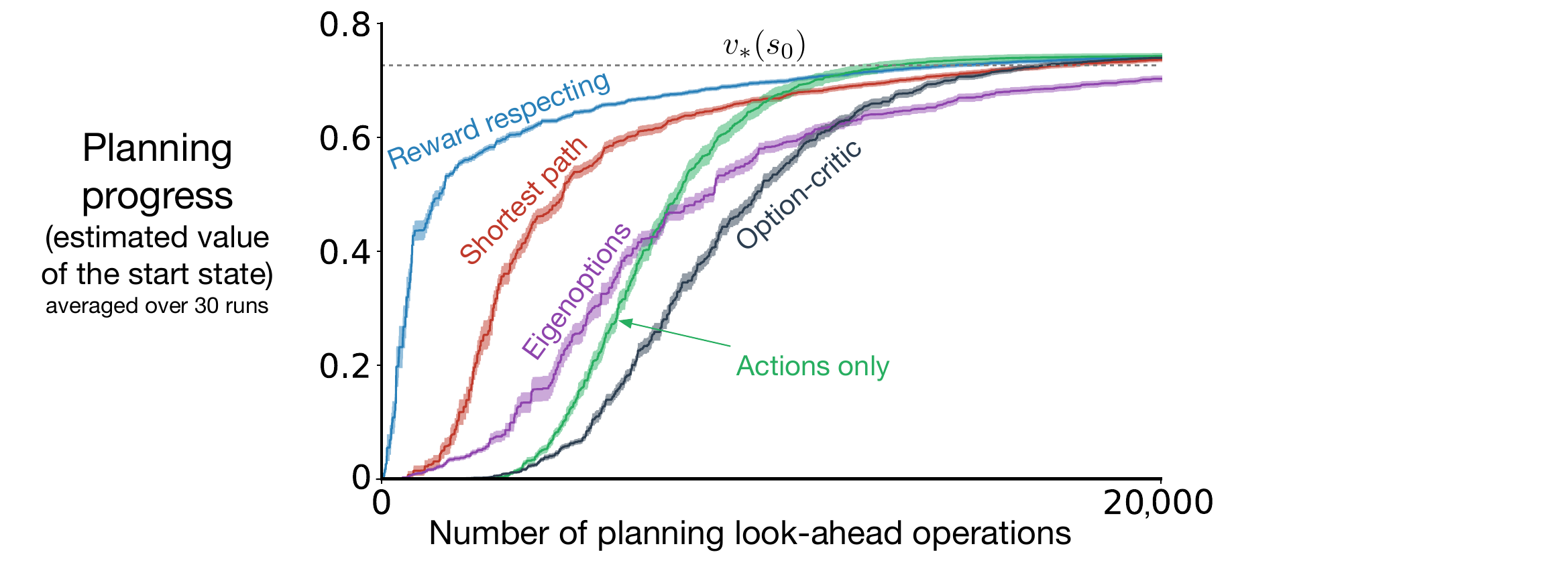}
    \caption{\textbf{Progress of planning} in the four-room gridworld with the models of options discovered in different ways.}
    \label{fig:planning-comparison}
\end{figure}

Given options created in the above four ways and the options corresponding to the primitive actions, we computed their models and conducted planning as described earlier in this paper. Detail on the progression of model learning for the reward-respecting options is provided in Appendix A. As a measure of the progress in planning, we recorded the approximate value of the start state after each step of AVI \eqrf{AVI}. Figure \ref{fig:planning-comparison} presents these planning results.

Planning with models of reward-respecting options was the fastest, followed by shortest-path options, eigenoptions, and option-critic options. Planning with models of the actions only was initially slower than planning with models of the eigenoptions, but then became faster and ultimately slightly surpassed all the other methods by this performance measure. The asymptotic superiority of the actions-only case may not be significant; remember that these are estimated values and not actual ones. An example of the difference is that the estimated values for many of the methods slightly exceed the largest possible actual value, $v_*(s_0)$, presumably due to maximization bias (van Hasselt, 2010, 2011) due to the stochastic environment. The actual values can be estimated directly by averaging Monte Carlo returns, but then the data is noisier and it is harder to see the differences between methods (see Appendix B).
The relatively poor performance of the option-critic approach may be surprising given that it only considers the main-task rewards. See Appendix C for a closer look suggesting that these options were poorly learned in much of the state space.

In the four-room gridworld, reward-respecting options seem more appropriate than alternative subtask formulations when these options are to be used in the STOMP progression. The advantages we saw in the two-rooms gridworld were retained as we extended to the larger problem, to multiple subtasks and options and to stochastic dynamics.


\section{Conclusions and future work}

In this paper we have further developed four ideas none of which is entirely new.
The most important is the idea of subtasks that respect the rewards of the original problem, both in their cumulants and in taking into account the estimated value of the state stopped in. Reward-respecting subtasks are a tiny subset of all possible subtasks, but they may be the most important in model-based reinforcement learning. The second idea is that of the STOMP progression for the development of temporally-abstract cognitive structure, an old idea which we have illustrated in greater detail and generality than in previous works. The third idea, arguably both the most novel and the most unproved, is that of subtasks for attaining state features. These are an appealing subset of reward-respecting subtasks because they achieve something pertaining to a state without requiring that states be completely observable. Achieving a state feature may be a part of effective plans, and feature attainment reduces the option discovery problem to that of deciding what features to maximize. Finally, the fourth idea developed in this paper is that of structuring learning algorithms in terms of a generic TD error \eqrf{delta} and a generic update procedure (\UWT).


There are several important areas in which future work could extend that presented here.
One natural next step would be to demonstrate all the steps of the STOMP progression operating simultaneously rather than sequentially as we have done them here. This appears straightforward, but doubtless some new issues will arise. Another natural extension would be to general (not one-hot) linear function approximation and multi-step eligibility traces. Although the algorithmic 
machinery described here applies immediately to both cases, the functionality was not exercised in this paper's experiments. 
An obvious further extension would be to use deep-learning neural networks for non-linear function approximation. However, a better idea might be to keep linear function approximation and instead extend the feature representation, which would retain the advantageous combination of linear function approximation and expectation models. A related issue is feedback from the later stages of the progression to earlier. For example, the effectiveness of planning as judged by search control methods (yet to be invented) should influence which models are used in planning and which features are used to form feature-attainment options and their models. Really, the feature construction step that precedes STOMP should also ultimately be influenced by the utility of features in all stages of learning and planning. This extension might be called FCSTOMP, or the ``Oak" architecture (Options And Knowledge, Sutton, Bowling \& Pilarski, 2022).

Here we have used planning only to improve the value function for the main task. A natural non-obvious extension would be to apply planning to improve the value functions for the subtasks.
Extending still further, everything in the form of a general value function should arguably be capable of being planned as well as learned. Because we have formulated the transition model of the environment as a collection of GVFs, it is intriguing to think that the model itself could be planned. That is, the higher-level parts of the transition model could be planned from the lower-level parts. The model of any option, for example, could be formed from planning with models of the primitive actions. This kind of planning---reaching a conclusion about how the world works from lower-level knowledge of how it works---seems deserving of the name \emph{reasoning}.


\section*{Acknowledgment}
The authors thank Joseph Modayil, Martha White, Michael Bowling, and Mark Ring for useful discussions, the anonymous reviewers for helping to clarify the papers contributions, and particularly Martin Klissarov for assistance analyzing our option-critic results and
Francesco Visin for his insightful feedback on an earlier draft. We also thank John Aslanides for the software producing option-critic options.

\section*{References}

\parskip=5pt
\parindent=0pt
\def\hangin{\hangindent=0.2in}
\def\bibitem#1{\hangin}

    \bibitem{Bacon} 
    Bacon, P.-L., Harb, J., Precup, D. (2017). The option-critic architecture. In \emph{Proceedings of the Association for Advancement of Artificial Intelligence}.
    
    \hangin Baranes, A., Oudeyer, P.-Y. (2013). Active learning of inverse models with intrinsically motivated goal exploration in robots. \textit{Robotics and Autonomous Systems 61}(1):49--73.
    
    \bibitem{Deisenroth} 
    Deisenroth, M. P., Rasmussen, C. E. (2011). PILCO: A model-based and data-efficient approach to policy search. In \emph{Proceedings of the International Conference on Machine Learning}.

    \hangin 
    Eysenbach, B., Gupta, A., Ibarz, J., Levine, S. (2019). Diversity is all you need: Learning skills without a reward function. In \textit{Proceedings of the International Conference on Learning Representations}.
    
    \bibitem{Harb18} 
    Harb, J., Bacon, P.-L., Klissarov, M., Precup, D. (2018). When waiting is not an option: Learning options with a deliberation cost. In \emph{Proceedings of the Association for Advancement of Artificial Intelligence}.
    
    \bibitem{Hochreiter} 
    Hochreiter, S., Schmidhuber, J. (1997). Long short-term memory. \emph{Neural Computation 9}(8):1735--1780.
    
    \bibitem{Jaderberg} 
    Jaderberg, M., Mnih, V., Czarnecki, W. M., Schaul, T., Leibo, J. Z., Silver, D., Kavukcuoglu, K. (2017). Reinforcement learning with unsupervised auxiliary tasks. In \textit{Proceedings of the International Conference on Learning Representations}.
    
    \hangin
    Jaeger, H. (2000). Observable operator models for discrete stochastic time series. \textit{Neural Computation 12}(6):1371--1398.
    
    \hangin
    Javed, K., Sutton, R. S. (in preparation). The big world hypothesis and the necessity of online continual learning in big worlds.
    
    \hangin
    Knoblock, C. A. (1994). Automatically generating abstractions for planning. \textit{Artificial Intelligence, 68}(2):243--302.
    
\hangin
Konda, V. R., Tsitsiklis, J. N. (2003). On actor-critic algorithms. \emph{SIAM Journal on Control and Optimization, 42}(4):1143--1166.

    \hangin
    Konidaris, G., Kaelbling, L. P., Lozano-Perez, T. (2018). From skills to symbols: Learning symbolic representations for abstract high-level planning. \textit{Journal of Artificial Intelligence Research 61}:215--289.

    \hangin
    Kudashkina, K. (2022). \textit{Model-based Reinforcement Learning Methods for Developing Intelligent Assistants} (PhD dissertation). Department of Engineering, University of Guelph.
    
    \bibitem{Lecun} 
    LeCun, Y., Bottou, L., Bengio, Y., Haffner, P. (1998). Gradient-based learning applied to document recognition. \emph{Proceedings of the IEEE 86}(11):2278--2324.

    \hangin
    Littman, M., Sutton, R. S., Singh, S. (2002). Predictive representations of state. \textit{Advances in Neural Information Processing Systems 14}.
    
    \bibitem{Machado23} 
    Machado, M. C., Barreto, A., Precup, D, Bowling, M. (2023). Temporal abstraction in reinforcement learning with the successor representation. \textit{Journal of Machine Learning Research 24}:1-69.

    \bibitem{Machado17} 
    Machado, M. C., Bellemare, M. G., Bowling, M. (2017). A Laplacian framework for option discovery in reinforcement learning. In \textit{Proceedings of the International Conference on Machine Learning}.
    
    \bibitem{Machado18} 
    Machado, M. C., Rosenbaum, C., Guo, X., Liu, M., Tesauro, G., Campbell, M. (2018). Eigenoption discovery through the deep successor representation. In \textit{Proceedings of the International Conference on Learning Representations}.
    
    \bibitem{McGovern} 
    McGovern, A., Barto, A. G. (2001). Automatic discovery of subgoals in reinforcement learning using diverse density. In \textit{Proceedings of the International Conference on Machine Learning}.
    
    \bibitem{Mnih} 
    Mnih, V., Kavukcuoglu, K., Silver, D., Rusu, A. A., Veness, J., Bellemare, M. G., \ldots Hassabis, D. (2015). Human-level control through deep reinforcement learning. \textit{Nature 518}:529--533.

    \hangin
Modayil, J., White, A., Sutton, R. S. (2014). Multi-timescale nexting in a reinforcement learning robot. \emph{Adaptive Behavior 22}(2):146--160.

\hangin
Ring, M.~B. (2021).
Representing knowledge as forecasts (and state as knowledge).
ArXiv:2112.06336.

    \hangin
    Rivest, R. L., Schapire, R. E. (1994). Diversity-based inference of finite automata. \textit{Journal of the ACM 41}(3):555--589.
    
    \hangin
    Sacerdoti, E. D. (1974). Planning in a hierarchy of abstraction spaces. \textit{Artificial Intelligence 5}(2):115--135.
    
    \bibitem{Silver} 
    Silver, D., Ciosek, K. (2012). Compositional planning using optimal option models. In \textit{Proceedings of the International Conference on Machine Learning}.
    
    \bibitem{Simsek04} 
    Simsek, \"O., Barto, A. G. (2004). Using relative novelty to identify useful temporal abstractions in reinforcement learning. In \textit{Proceedings of the International Conference on Machine Learning}.
    
    \bibitem{Simsek05} 
    Simsek, \"O., Wolfe, A. P., Barto, A. G. (2005). Identifying useful subgoals in reinforcement learning by local graph partitioning. In \textit{Proceedings of the International Conference on Machine Learning}.
    
    \bibitem{Singh} 
    Singh, S. P., Barto, A. G., Chentanez, N. (2004). Intrinsically motivated reinforcement learning. In \textit{Advances in Neural Information Processing Systems}.
    
    \bibitem{Solway} 
    Solway, A., Diuk, C., Cordova, N., Yee, D., Barto, A. G., Niv, Y., Botvinick, M. (2014). Optimal behavioral hierarchy. \textit{PLoS Computational Biology 10}(8).

    \bibitem{Sorg} 
    Sorg, J., Singh, S. P. (2010). Linear options. In \textit{Proceedings of the International Conference on Autonomous Agents and Multiagent Systems}.
    
    \bibitem{Sutton84} 
    Sutton, R. S. (1984). \textit{Temporal Credit Assignment in Reinforcement Learning} (PhD dissertation). Department of Computer Science, University of Massachusetts, Amherst.
    
    \bibitem{Sutton88} 
    Sutton, R. S. (1988). Learning to predict by the methods of temporal differences. \textit{Machine Learning 3}:9--44 (important erratum p.~377).

    \bibitem{Sutton95}  
    Sutton, R. S. (1995). TD models: Modeling the world at a mixture of time scales. In \textit{Proceedings of the International Conference on Machine Learning}.
    
    \bibitem{Sutton18} 
    Sutton, R. S., Barto, A. (2018). \textit{Reinforcement Learning: An Introduction} (2nd ed.). MIT Press.
    
    \hangin
    Sutton, R. S., Bowling, M., Pilarski, P. M. (2022). The Alberta plan for AI research. ArXiv:2208.11173.
    
\hangin
Sutton, R. S., McAllester, D. A., Singh, S. P., Mansour, Y. (2000). Policy gradient methods for reinforcement learning with function approximation. In \emph{Advances in Neural Information Processing Systems 12}, pp.~1057--1063. MIT Press, Cambridge, MA.

    \bibitem{Sutton11} 
    Sutton, R. S., Modayil, J., Delp, M., Degris, T., Pilarski, P. M., White, A., Precup, D. (2011). Horde: A scalable real-time architecture for learning knowledge from unsupervised sensorimotor interaction. In \textit{Proceedings of the International Conference on Autonomous Agents and Multiagent Systems}.
    
    \bibitem{Sutton99} Sutton, R. S., Precup, D., Singh, S. (1999). Between MDPs and semi-MDPs: A framework for temporal abstraction in reinforcement learning. \textit{Artificial Intelligence 112}(1--2):181--211.
    
\hangin
van Hasselt, H. (2010). Double Q-learning. In \emph{Advances in Neural Information Processing Systems 23}, pp.~2613--2621.  Curran Associates, Inc.

\hangin
van Hasselt, H. (2011). \emph{Insights in Reinforcement Learning: Formal Analysis and Empirical Evaluation of Temporal-difference Learning}. SIKS dissertation series number 2011-04.

\hangin
Veeriah, V. (2022). \emph{Discovery in Reinforcement Learning} (Doctoral dissertation). Department of Computer Science and Engineering, University of Michigan. See Chapter 7.

    \bibitem{Wan} Wan, Y., Zaheer, M., White, A., White, M., Sutton, R. S. (2019). Planning with expectation models. In \textit{Proceedings of the International Joint Conference on Artificial Intelligence}.

\clearpage

\appendix

\newpage
\section{Additional results in the four-room gridworld}\label{sec:exp3}
\setcounter{figure}{0}

\begin{figure*}[ht]
    \centering
     \includegraphics[width=1.05\textwidth]{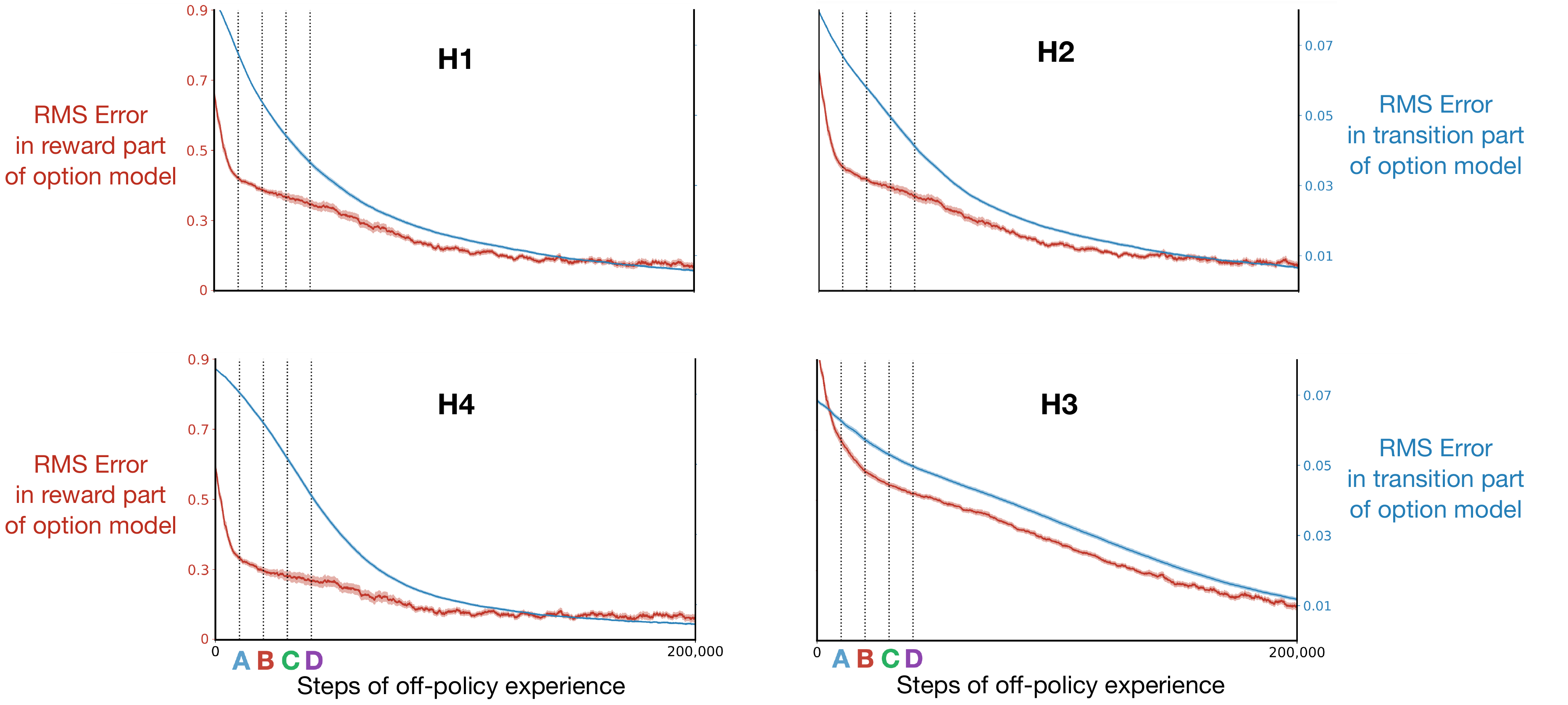}
     \caption{\textbf{Model learning in the four-room gridworld}. The time course of learning of the transition parts (blue, right scale) and reward parts (red, left scale) of the models of each of the four options. In all cases the error becomes dramatically smaller, but here the error will never converge to zero because of the stochasticity of the environmental dynamics. All lines were averaged over 30 runs and the shading represents the standard~error.}
     \label{fig:4model-learning}
\end{figure*}

\begin{figure*}[hb!]
\hspace*{1cm}
     \includegraphics[width=0.75\textwidth]{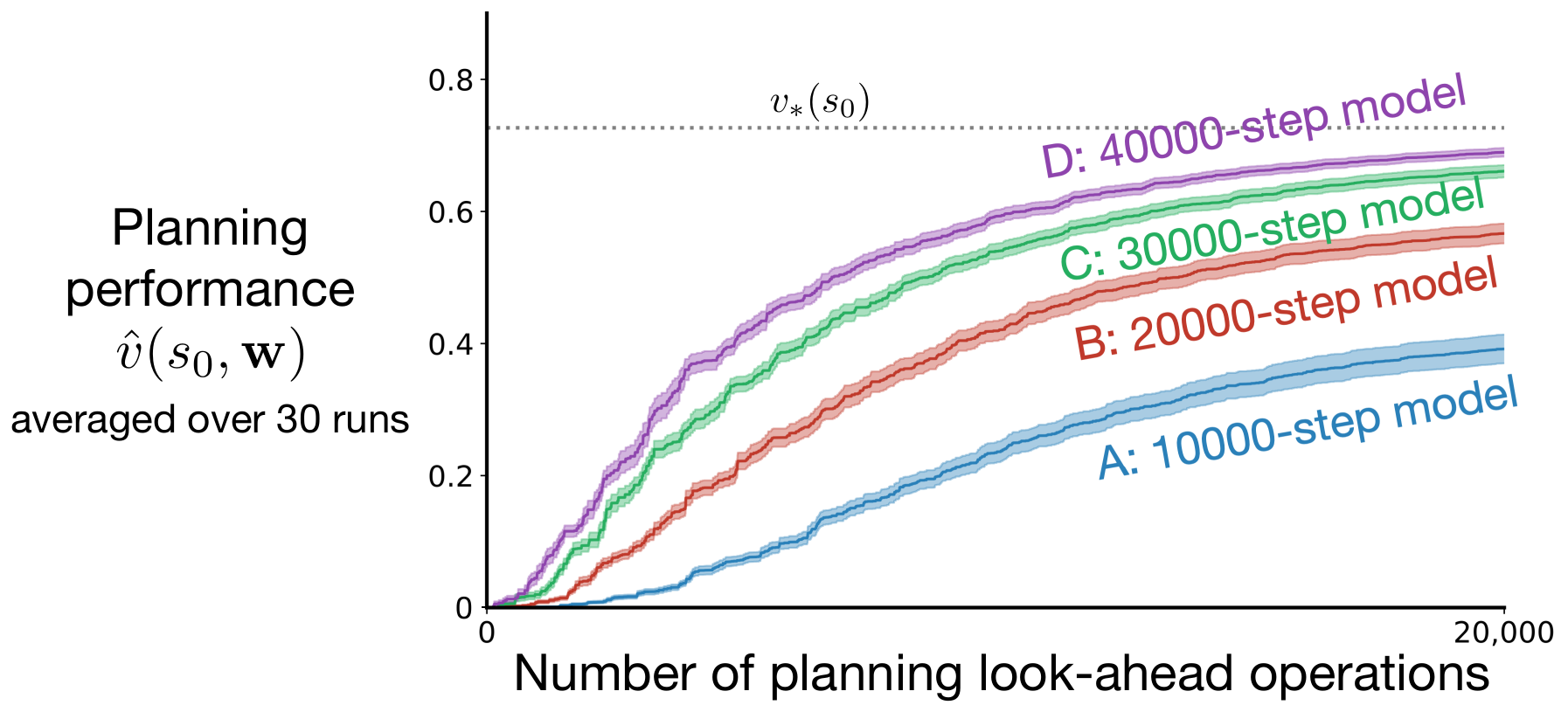}
     \caption{\textbf{Performance of planning} for learned, approximate models with different amounts of training, corresponding to the vertical lines in Figure~\ref{fig:4model-learning}.}
     \label{fig:planning-appendix}
\end{figure*}

In Section~\ref{sec:final_experiments} we presented results with the STOMP progression using four reward-respecting feature-attainment subtasks. In this appendix we present additional results pertaining to model learning (Figure~\ref{fig:4model-learning}) and to planning with the imperfectly learned models (Figure \ref{fig:planning-appendix}).

%

%
%
\newpage
\section{Monte Carlo estimates of planning performance}
\setcounter{figure}{0}

Throughout the paper we have presented planning results in which the \emph{estimated} value of the start state, $\hat v(s_0, \w)$, was used as a proxy for the value of the policy $\pi_\w$ induced by the state values. However, in some cases $\hat v(s_0, \w)$ is a poor proxy for $v_{\pi_\w}(s_0)$. An example of the discrepancy is that in early stages of planning the value estimates will always be near zero because \w\ starts near zero and $\hat v$ is linear in \w, but $v_{\pi_\w}(s_0)$ could be positive or negative depending on the environment. Specifically, in our initial illustrative example, the two-room gridworld with the field of $-1$s, the initial returns will have $-1$s in them and overall will probably be negative, whereas the estimated values in the initial stages of planning (see Figure 1) are slightly positive.

In this appendix we redo all the paper's planning results with a more direct Monte-Carlo estimate of $v_{\pi_\w}(s_0)$. 

First we need a clear specification of $\pi_\w$. 
Let $g(s)$ denote the greedy option in state $s$ given the current state-value weight vector for the main task, $\w$, and the model, $\r$ and $\n$:
\begin{equation}
g(s) \doteq \arg\max_{o\in\O} \bigl[\r(\x(s),o) + \v(\n(x(s),o),\w)\bigr], 
~~~~~~\forall s\in\S.
\end{equation}
For states $s$ in which $g(s)$ is an \emph{action}, $\pi_\w$ was defined to take that action deterministically. For states in which $g(s)$ was an \emph{option}, $\pi_\w$ was defined as a  stochastic selection from the actions with probabilities given by the soft-max policy for that option \eqrf{softmax}.

\begin{figure}
\hspace*{0.6cm}
\includegraphics[width=0.83\textwidth]{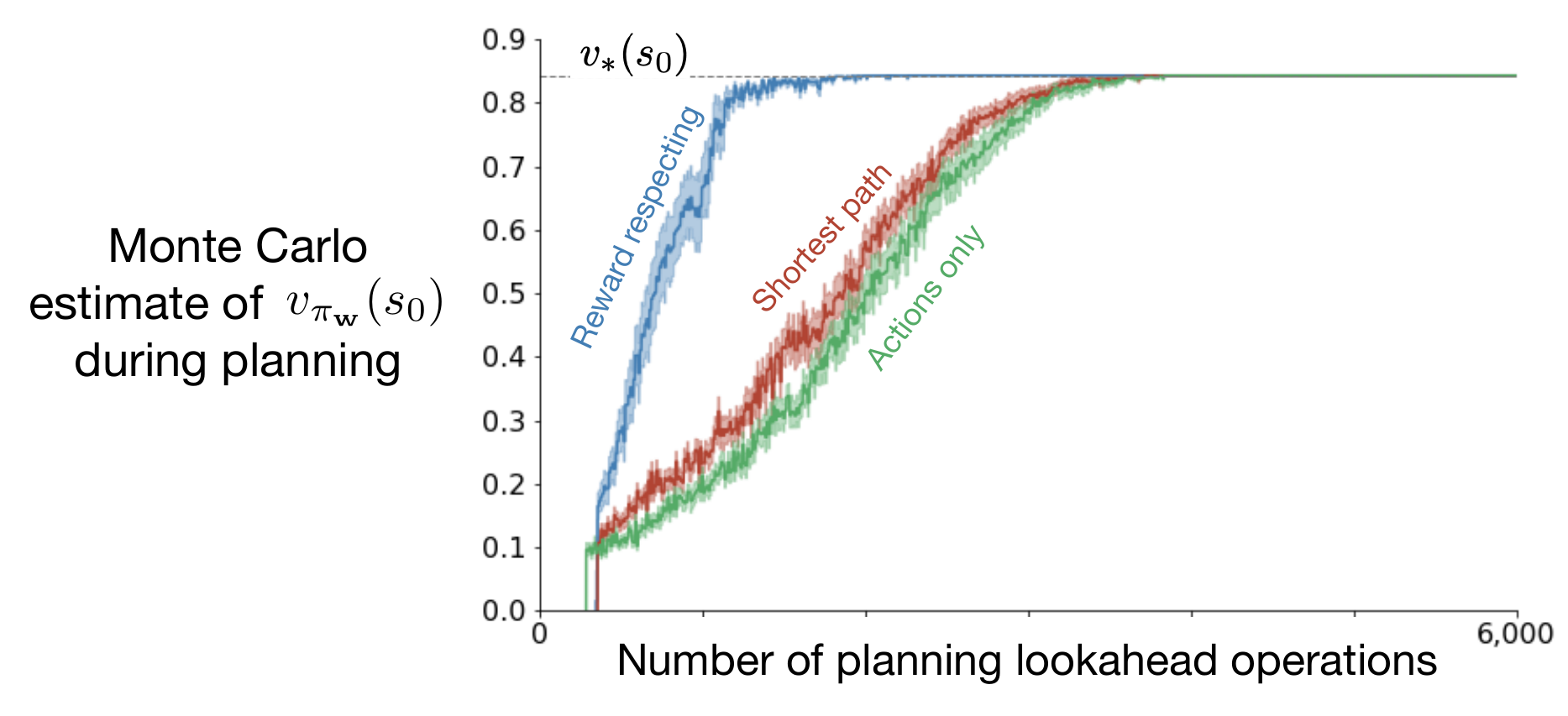}
\caption{\label{fig:fig1_mc}
\textbf{Same experiment as Figure \ref{fig:exp1}}, but with Monte Carlo estimate of value. }
\end{figure}

\begin{figure}
\hspace*{0.6cm}
\includegraphics[width=0.83\textwidth]{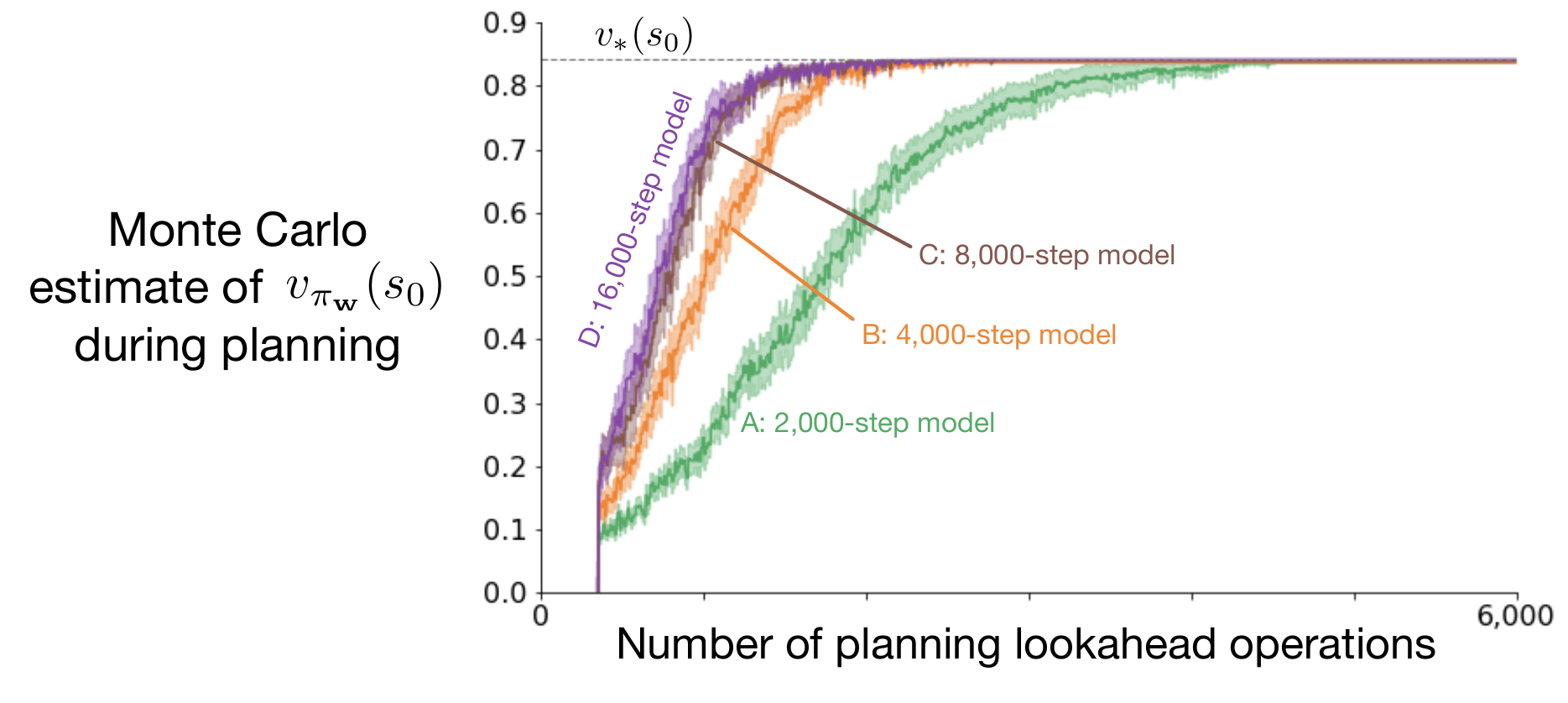}
\caption{\label{fig:fig3_mc}
\textbf{Same experiment as inset in Figure 3}, but with Monte Carlo estimate of value. }
\end{figure}

During planning, after each update of \w\ by \eqrf{AVI}, one trajectory from start state to termination was generated by following $\pi_\w$ with the real environmental dynamics. The return on that trajectory was recorded as a noisy Monte Carlo estimate of $v_\w(s_0)$. If the trajectory did not terminate after 1000 steps, then the partial return was used as the estimate of the return ($\gamma^{1000}\approx 0.00004$). To reduce the noise, these estimates were averaged over 30 runs and then averaged over a bin of updates to produce the plots that follow. 
Figures \ref{fig:fig1_mc} and  \ref{fig:fig3_mc} used a bin size of 10, and the others used a bin size of 50.
The values for the first few hundred updates in all cases were negative and are not shown in the plots (they are clipped at zero).

\begin{figure}
\hspace*{0.6cm}
\includegraphics[width=0.83\textwidth]{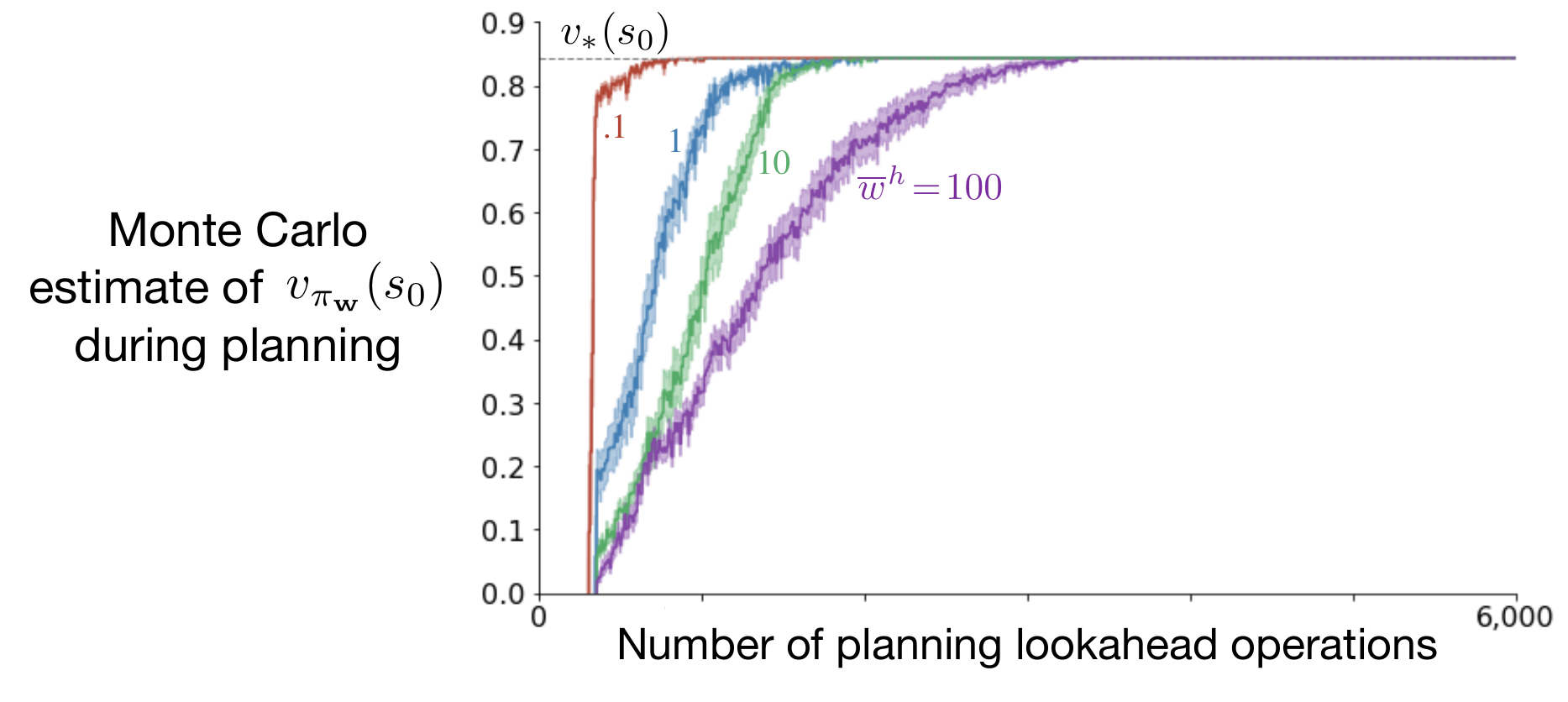}
\caption{\label{fig:fig4_mc}
\textbf{Same experiment as Figure 5}, but with Monte Carlo estimate of value. }
\end{figure}

\begin{figure}
\includegraphics[width=0.95\textwidth]{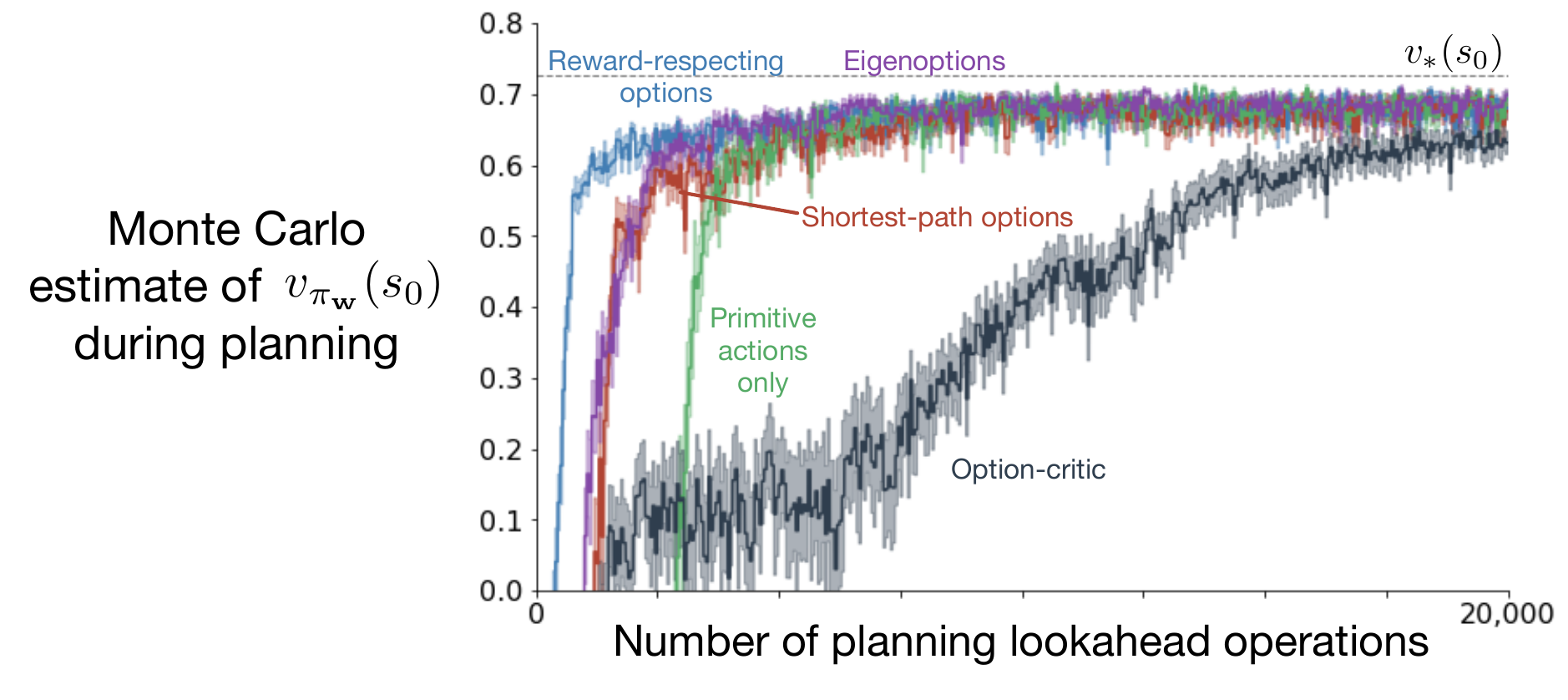}
\caption{\label{fig:mc-4room-comparison}
\textbf{Same experiment as Figure 8}, but with Monte Carlo estimate of value. }
\end{figure}

\begin{figure}
\includegraphics[width=0.95\textwidth]{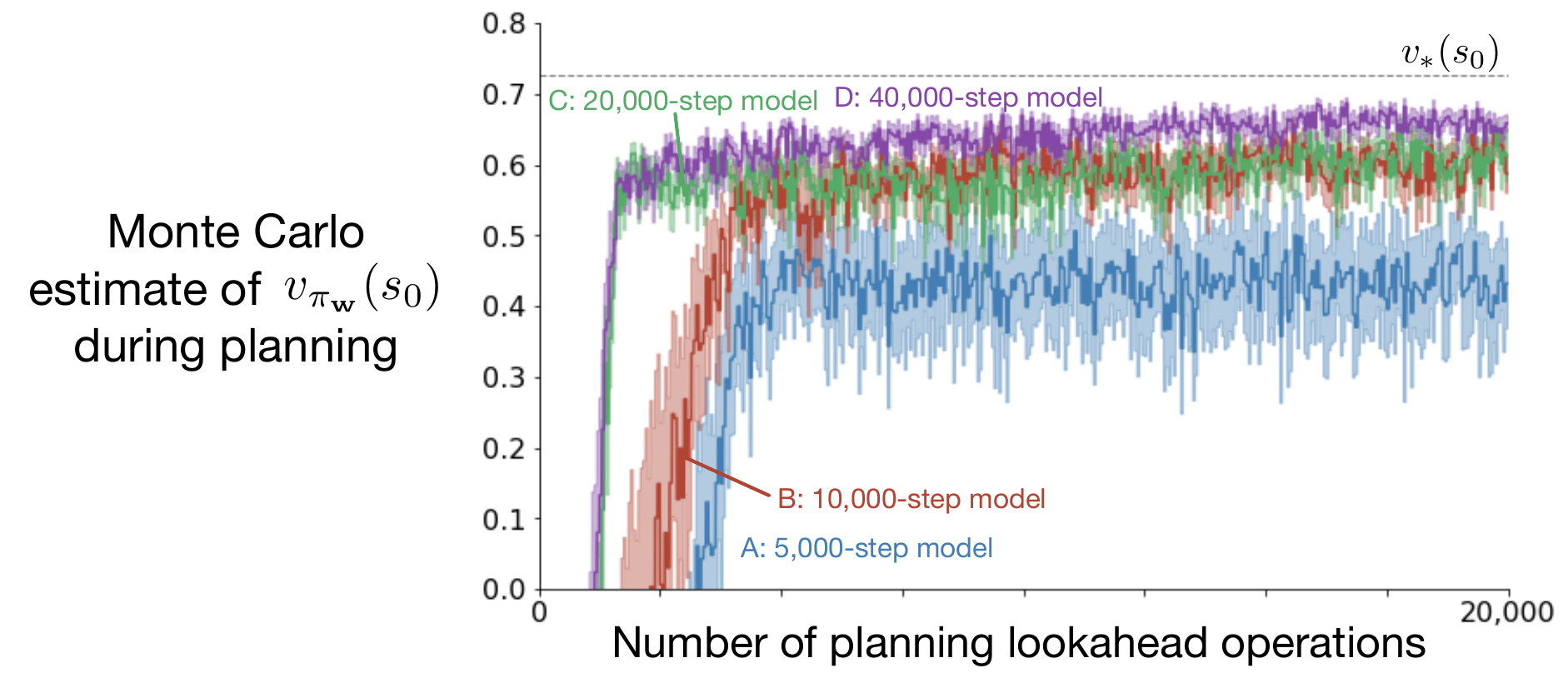}
\caption{\label{fig:figB9_mc}
\textbf{Same experiment as Figure \ref{fig:planning-appendix}}, but with Monte Carlo estimate of value. }
\end{figure}

\newpage
\section{Understanding the option-critic's performance}
\setcounter{figure}{0}

As discussed in the main text, the poor performance of planning with option-critic options (Figures~\ref{fig:planning-comparison} and~\ref{fig:mc-4room-comparison}) may be surprising at first because these are reward-respecting options. Further inspection highlights the importance of appropriately choosing the state distribution we learn options from, and the distribution we use to sample state-feature vectors when performing AVI.

\begin{figure*}[t]
\centering
\includegraphics[width=0.99\textwidth]{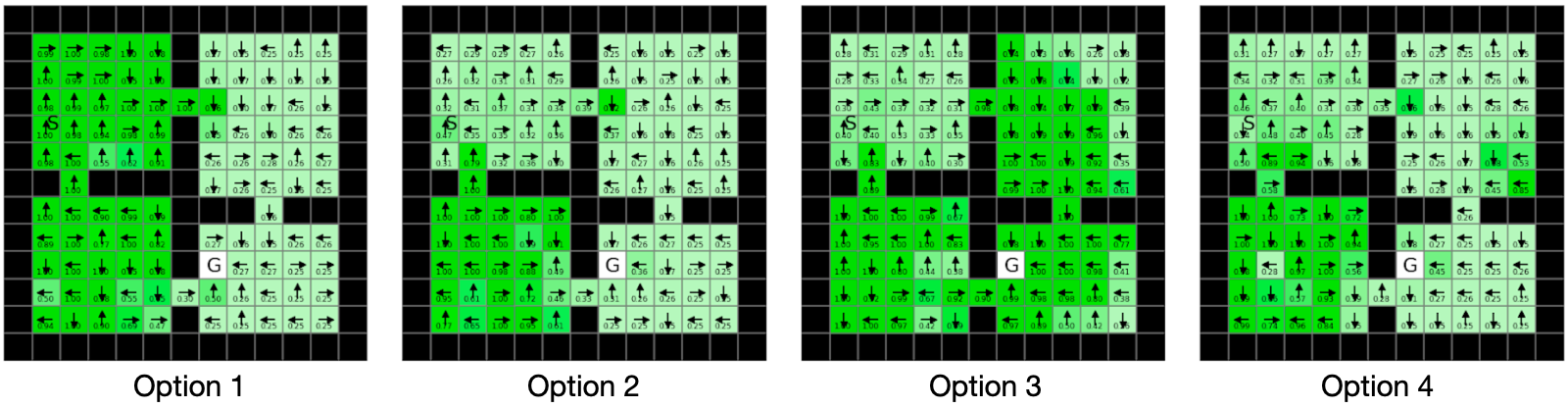}
\caption{\textbf{Option policies} learned by the option-critic method in a representative run. The arrows indicate the greedy action in each state, and the numbers indicate its probability of being taken (the actual policies were stochastic).
Darker green indicates states in which the policy is more deterministic.}
\label{fig:oc_option_policy_2}
\end{figure*}

\begin{figure*}[t]
\centering
 \includegraphics[width=0.99\textwidth]{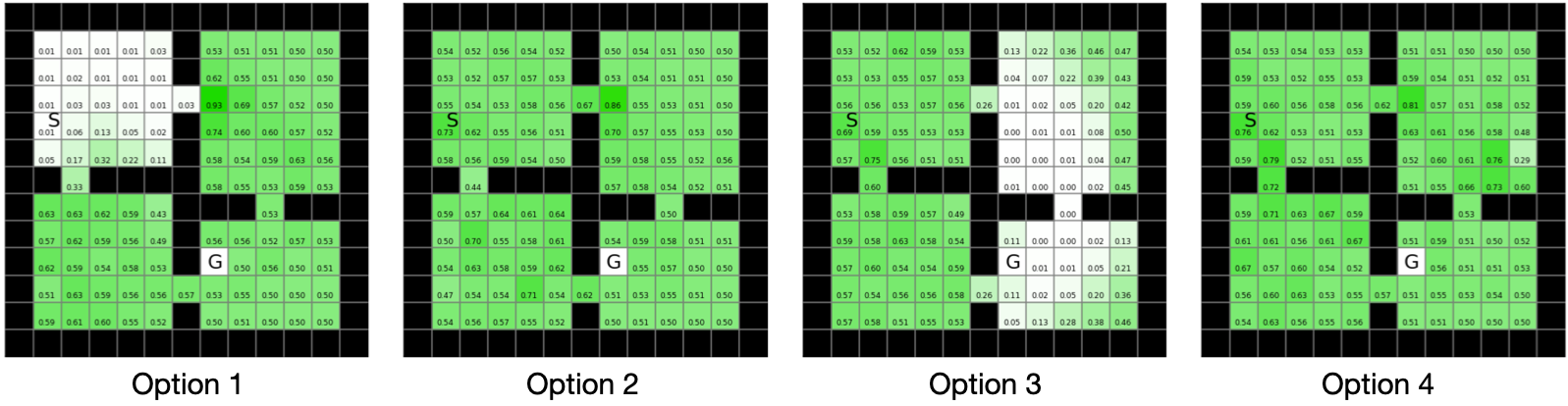}
\caption{\textbf{Stopping probabilities} for the option policies depicted above in Figure~\ref{fig:oc_option_policy_2}. Darker green indicates states in which the option stops more frequently. In this case, starting from the start state {\sf S}, Option 1 is unlikely to stop until reaching the hallway state, and then Option 3 is unlikely to stop until reaching the goal state.} 
\label{fig:oc_termination_2}
\end{figure*}

\begin{figure*}[b]
\centering
\vspace*{-.0in}
 \includegraphics[width=0.7\textwidth]{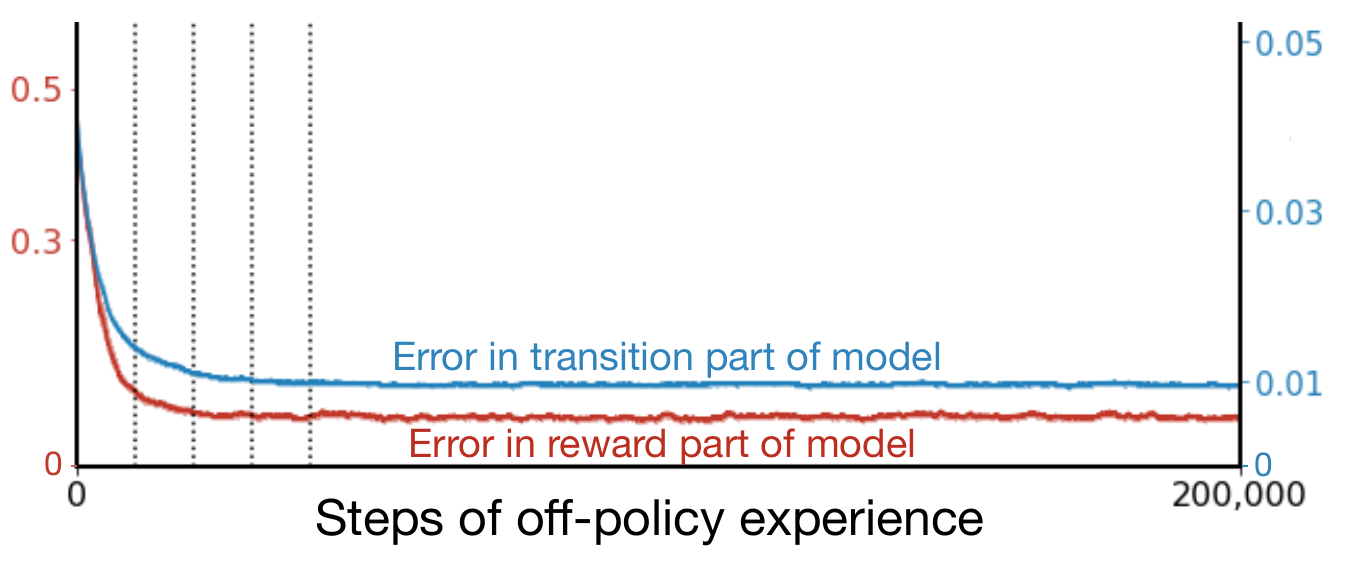}
\caption{\textbf{Model learning in the four-room gridworld} with an option-critic option. The time course of learning of the transition part (blue, right scale) and reward part (red, left scale) of the models of one of the options learned by the option-critic. All lines were averaged over 30 runs.}
\label{fig:model_learning_oc}
\end{figure*}

In the four-room gridworld, as expected, the option-critic learns near-optimal policies that consistently reach the goal state. However, because the option-critic is an \emph{on-policy} method, it does not learn accurate estimates of the option values across the whole state-space, but only across the set of states the options are likely to visit---see representative learned policies in  Figure~\ref{fig:oc_option_policy_2} and stopping probabilities in Figure~\ref{fig:oc_termination_2}. Because we are selecting state-feature vectors \x\ in a random sequence when performing AVI, there are several state-feature vectors \x\ with inaccurate values that hinder planning performance, even though the models we learn are very accurate (see Figure~\ref{fig:model_learning_oc}).

More effective techniques for selecting the states which will be updated, a problem known as \emph{search-control}, could make options learned by the option-critic more effective, but this is currently an open problem.

The options learned by the option-critic, shown in Figures~\ref{fig:oc_option_policy_2} and~\ref{fig:oc_termination_2}, highlight the importance of defining stopping values that are different from the estimated values of the state the option stops in. Because the option-critic does not do that, it rarely learns four distinct, useful options. Some of the options learned by the option-critic often look like random policies in most of the state space. Notice that different stopping values, such as deliberation cost (Harb et al., 2018), would not completely address this issue because of the on-policy nature of the underlying method. This discussion highlights the importance of taking into account how the option models will be used in planning.

\end{document}